\documentclass[utf8]{notfrontiersFPHY}

\usepackage[utf8]{inputenc}

\setcitestyle{square}
\usepackage{url,hyperref,lineno,microtype,subcaption}
\usepackage[onehalfspacing]{setspace}

\usepackage{tikz,amsmath,amssymb,bm,color}
\usetikzlibrary{shapes, shapes.geometric, arrows, positioning, calc,
  decorations.pathreplacing, matrix, external}

\usepackage{mathtools} 

\usepackage{upgreek} 

\usepackage{anyfontsize} 

\newcommand{\comment}[2]{\hspace{0in}#2} 
\newcommand{\overbar}[1]{\mkern 1.5mu\overline{\mkern-1.5mu#1\mkern-1.5mu}\mkern 1.5mu}

\def\keyFont{\fontsize{8}{11}\helveticabold }
\def\firstAuthorLast{Hennen {et~al.}}
\def\Authors{T. Hennen\,$^{1}$, A. Elias\,$^{2}$, J. F. Nodin\,$^{3}$,
  G. Molas\,$^{3,4}$, R. Waser\,$^{1}$, D. J. Wouters\,$^{1}$, and D. Bedau\,$^{2,*}$}
\def\Address{$^{1}$IWE II, RWTH Aachen University, Aachen, Germany \\
  $^{2}$Western Digital San Jose Research Center, CA, USA \\
  $^{3}$CEA, LETI, MINATEC Campus, Grenoble, France\\
$^{4}$Now with Weebit Nano Ltd.}

\def\corrEmail{daniel.bedau@wdc.com}

\def\x{\bm{x}}  
\def\xhat{\widehat{\bm{x}}} 
\def\R{\mathbb{R}}
\def\N{\mathcal{N}}
\def\G{\bm{\Gamma}}
\def\eps{\bm{\epsilon}}

\def\IHHRS{I_{\text{HHRS}}}
\def\ILLRS{I_{\text{LLRS}}}
\def\IHRSn{I_{\text{HRS,n}}}
\def\ILRSn{I_{\text{LRS,n}}}
\def\IHRSnp1{I_{\text{HRS,n+1}}}
\def\ILRSnp1{I_{\text{LRS,n+1}}}
\def\Umax{U_{\text{max}}}

\newcommand{\inv}{^{\raisebox{.2ex}{$\scriptscriptstyle\text{-}1$}}}

\begin{document}
\onecolumn
\firstpage{1}

\title[Fast Stochastic Synapse Model]{A High Throughput Generative Vector Autoregression Model for Stochastic Synapses}

\author[\firstAuthorLast ]{\Authors} 
\address{} 
\correspondance{} 

\extraAuth{}

\maketitle

\begin{abstract}
\section{}
By imitating the synaptic connectivity and plasticity of the brain, emerging
electronic nanodevices offer new opportunities as the building blocks of
neuromorphic systems. One challenge for large-scale simulations of computational
architectures based on emerging devices is to accurately capture device
response, hysteresis, noise, and the covariance structure in the temporal domain as
well as between the different device parameters. We address this challenge with
a high throughput generative model for synaptic arrays that is based on a
recently available type of electrical measurement data for resistive memory
cells. We map this real world data onto a vector autoregressive stochastic
process to accurately reproduce the device parameters and their
cross-correlation structure. While closely matching the measured data, our model
is still very fast; we provide parallelized
implementations for both CPUs and GPUs and demonstrate array sizes above one
billion cells and throughputs exceeding one hundred million weight updates per
second, above the pixel rate of a 30 frames/s 4K video stream.
\tiny
 \keyFont{ \section{Keywords:} Neuromorphic Computing, Machine Learning, Emerging
   Technologies, Stochastic Model, Synapse, ReRAM, Julia, GPU}
\end{abstract}

\section{Introduction} 

Recent trends in computing hardware have placed increasing emphasis on
neuromorphic architectures implementing machine learning (ML) algorithms
directly in hardware. Such bio-inspired approaches, through in-memory
computation and massive parallelism, excel in new classes of computational
problems and offer promising advantages with
respect to power consumption error resiliency.
While CMOS-based neuromorphic computing (NC) implementations have made
substantial progress recently, new materials and physical mechanisms may
ultimately provide better opportunities
for energy efficiency and scaling \cite{burr_neuromorphic_2017, sangwan_neuromorphic_2020, milo_memristive_2020}.

\begin{figure}[h]
\begin{center}
  \includegraphics[]{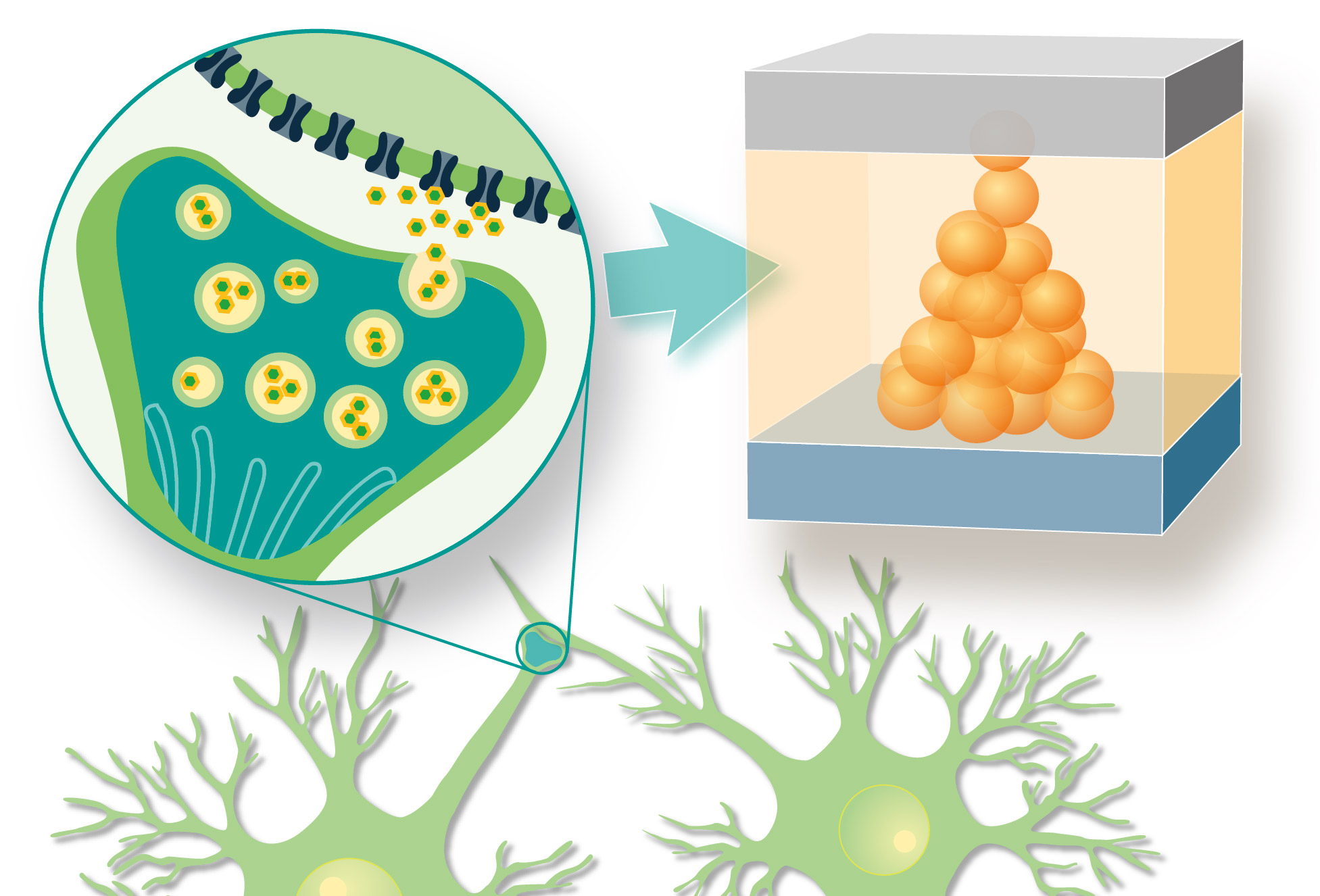}
\end{center}
\caption{
  In analogy to biological synapses, two terminal solid state nanodevices such
as ReRAM can store synaptic weights as electrical resistance states. The
devices, consisting simply of patterned metal-insulator-metal material stacks,
have an adjustable resistance level determined by the ionic configuration inside
the insulating layer. This nano-ionic mechanism also exhibits non-ideal
properties such as stochasticity and noise.
}\label{fig:synapse}
\end{figure}

A specific functionality required in NC applications is the ability to mimic
synaptic connections and plasticity by allowing the storage of large numbers of
interconnected and continuously adaptable resistance values.
Several candidate memory technologies such as MRAM, ReRAM, PCM, CeRAM, are
emerging to cover this behavior using different physical mechanisms
\cite{chen_emerging_2014, yu_emerging_2016, you_zhou_mott_2015, liu_dynamics_2014}.
Among these, ReRAM is attractive for its simplicity of materials and device
structure, providing the necessary CMOS compatibility and scalability \cite{waser_redox-based_2009}.
ReRAM is essentially a two terminal nanoscale electrochemical cell, whose
variable resistance state is based on manipulation of the point defect
configuration in the oxide material (depicted in Fig. \ref{fig:synapse}). This
redox-based switching mechanism is intrinsically analog, allowing stable
resistance levels to be stored and adjusted through application of bipolar
voltage stimuli. However, non-idealities such as stochasticity, nonlinearity,
and noise are prominent features of these devices that critically impact the
performance of neuromorphic systems composed of them \cite{kim_impact_2018}.

Modern ML models have reached an astonishingly large and ever-increasing size,
with recent examples exceeding a hundred billion weights
\cite{brown_language_2020}. Before comparable neuromorphic hardware using
artificial solid-state synapses can become a reality, large-scale network
designs need to first be implemented and evaluated in computer simulations.
Training, validation, and optimization of such networks is a process that
involves a huge number of simulated devices, voltage pulses, and current
readouts. Within this process, it is important to accurately consider the
constraints of the underlying hardware in detail. Therefore, lightweight, fast,
and accurate stochastic simulations of the individual
synaptic devices is a key requirement.

Traditionally, device modelling begins with a
physical description of the materials and processes involved. In the
case of ReRAM, the physical situation is immensely complicated with many degrees
of freedom, and accurate modelling is a wide-scale and ongoing research
undertaking. Efforts in this direction are
motivated by advancing an understanding of physical and chemical dependencies
that can in principle inform design choices on physically justified grounds. In
the past decade, many different computational techniques have been employed to
furnish device models, from ab initio density-functional theory (DFT), molecular
dynamics (MD), kinetic Monte Carlo (KMC), finite element method (FEM), as well as
ordinary differential equation (ODE) and differential algebraic equation (DAE)
solvers \cite{jiang_using_2017, stewart_diffusion_2019,
kopperberg_consistent_2021, ascoli_art_2015, messaris_data-driven_2018}. The
resulting models exist on a spectrum of physical
abstraction, such that the cost of increasing computational speed is generally a
trade-off in physical accuracy/detail \cite{ielmini_physics-based_2017}.

Device models that naturally encompass stochasticity do so at the cost of
complexity needed to compute the physical scenario in high detail. For example,
atomistic KMC simulates switching processes with atomic precision and is
inherently stochastic but requires hours of computation per cycle even for small
individual cell volumes (e.g. 125~nm$^2$ \cite{Abbaspour_studying_2020}). At the
other end of the spectrum, dynamic models based on numerical solutions of
ODE systems are designed to run significantly faster while sometimes aiming to
remain physically realistic. However, their higher speed invariably comes at the
cost of approximations, simplifications, and
omissions of physical reality. Typically, device operation is distilled to a
dynamical description of one or two state variables, such as a conducting
filament length, radius, or a defect concentration.

Due in part to ambiguity in their high dimensional parameter space, a given
ODE model encompasses a diverse range of possible cell behaviors and has the
flexibility to approximately match measurement
data \cite{reuben_modeling_2019, mayer_drawing_2010}. However, fitting the 
model to data is commonly an ad-hoc, manual, and/or unspecified procedure.
Having dispensed with the atomistic sources of
variability, ODE models are fully deterministic by default.
Where stochasticity is required, it is accounted for by
injecting noise into the state variables or parameters of the model
\cite{li_device_2017, maria_puglisi_bipolar_2015,
bengel_variability-aware_2020}. Due to the unique experimental challenges
posed by electrical measurement of ReRAM, the data used for fitting is not
necessarily statistically sufficient nor measured under relevant electrical
conditions and timescales. While models can be tuned by hand to roughly match
the dispersion observed in a measurement \cite{jiang_compact_2016,
chen_compact_2015}, they generally fail to accurately reproduce the complex
statistical properties of actual devices.

The main purpose of ODE device models is to be computationally efficient enough
to support circuit simulation. Still, nonlinear ODE solvers require many finely
spaced timesteps and a considerable amount of total time to compute dynamical
trajectories. Although they have been successfully used to demonstrate small
scale circuitry such as logic elements and small crossbar arrays
\cite{siemon_memristive_2019, bocquet_compact_2014, huang_compact_2017,
wald_understanding_2019}, benchmarks or indications of run time for ODE-based
simulations have so far not been supplied. With the exception of extremely
small ML model sizes on the order of $10^3$ weights or below,
demonstrations of network performance are expected to remain
computationally intractable via conventional circuit simulation.

In this article, we address these device modelling challenges with a new type of
generative model for arrays of artificial synapses. The main objective of the
model is to accurately reproduce the statistical
properties of fabricated devices while remaining computationally lightweight.
Starting with newly available electrical measurement data as an input, this
phenomenological model is systematically fit using a well defined statistical
regression analysis. The exclusive use of easily computable analytical
expressions provides close
quantitative agreement with relevant experimental observation. Taking advantage of
parallel resources on a modern CPU and GPU, we demonstrate the ability to
simulate hundreds of millions of synaptic connections with over $10^8$ weight
updates per second. With its high throughput
and low memory footprint, the model can be usefully employed
to simulate large arrays of solid-state synapses for investigation of
emerging NC concepts on a relevant scale.

\bigskip
\section{Method} 

The basic requirement for an electronic
device serving as an artificial synapse is to moderate the flow of electrical
signals through connections in a network. Left undisturbed, the device ideally
maintains a fixed weight, or dependence between the voltage across the two
device terminals, $U$, and the resulting current through the device, $I$.
Further, for learning there must be some means of affecting the weight in a
durable way. ReRAMs are bipolar devices that have an adjustable
(potentially nonlinear) non-volatile resistance state, which is based on the
size and shape of a conducting filament that partially or fully bridges the
insulating gap of the oxide material. Simplistically, when $U$ exceeds certain
threshold levels, the resistance state begins to transition toward lower or
higher values depending on the voltage polarity, which corresponds to growth and
shrinkage of the conducting filament. When the filament only partially bridges
the insulating gap, conduction may be limited for example by tunneling through a
Schottky barrier of a material interface, leading to a relatively high
resistance levels \cite{yang_memristive_2008, waser_redox-based_2009}. As the
filament grows and gradually bridges the gap, the resistance decreases as
conduction transitions into the ohmic type.

\begin{figure}[h]
\begin{center}
  \includegraphics[]{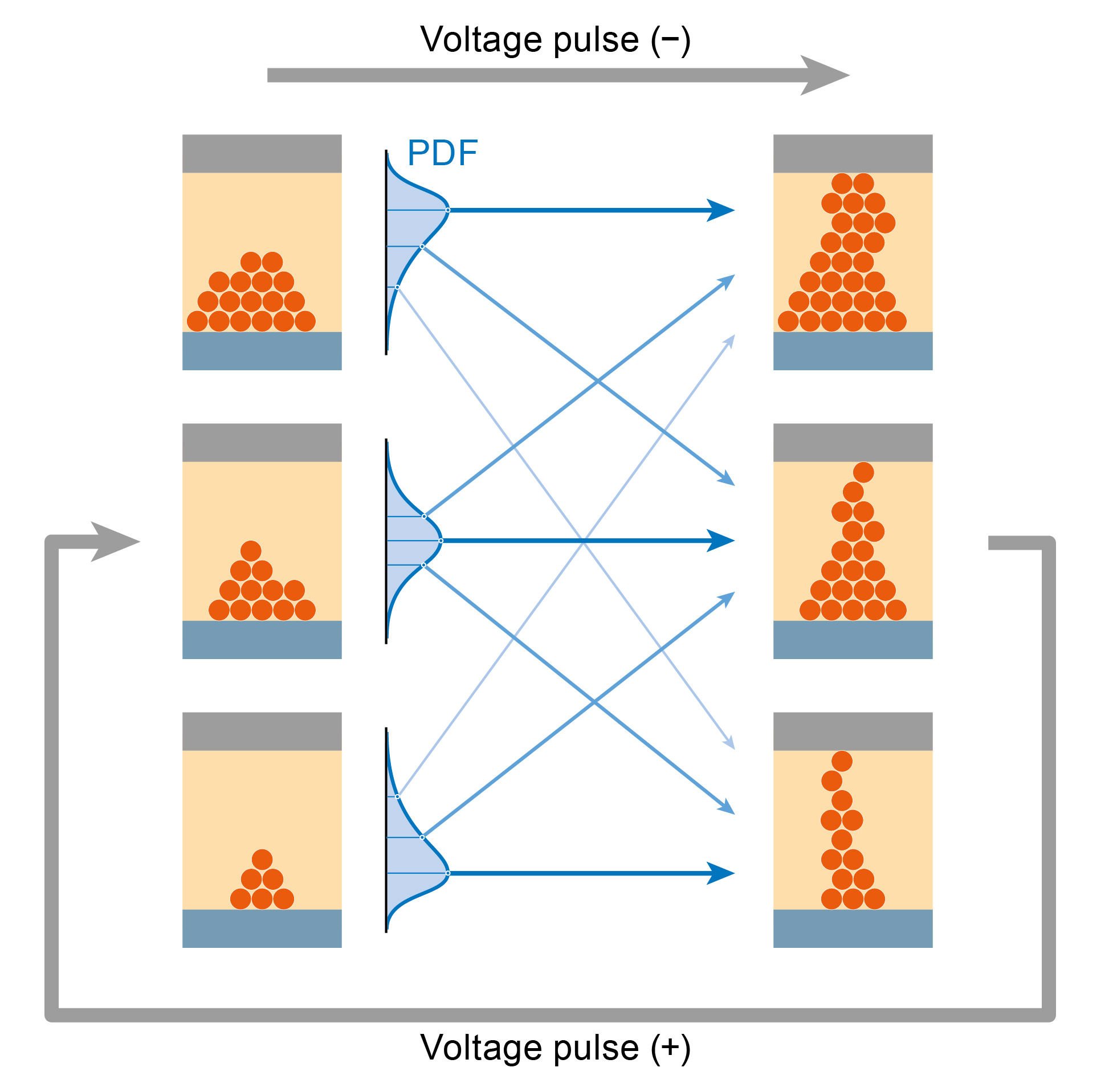}
\end{center}
\caption{
  Resistance states reached in a synaptic ReRAM device through application of
voltage pulses exhibit a probabilistic dependence on past states, leading to
long range correlations that also involve other parameters such as the voltage
thresholds required for switching. Starting with effectively
infinite state possibilities, represented by the three
cells on the left, an applied voltage pulse brings about a set of transition
probabilities to many possible future states (right).
}\label{fig:correlated_cartoon}
\end{figure}

In designing our model, we place high priority on speed and fitting accuracy.
One of the beginning assumptions is that in every possible device state, the
current can be represented by a linear mixture of two fixed polynomials in $U$.
These two polynomials, which are each estimated from a fit to measurement data,
can be thought of as limiting cases for the highest possible high resistance
state, $\IHHRS(U)$, and lowest possible low resistance state, $\ILLRS(U)$. The
device current in all possible resistance states is then given by
\begin{equation}\label{eq:poly_mixture}
I(r, U) = r \IHHRS(U) + (1-r) \ILLRS(U),
\end{equation}
conveniently reducing the description of the conduction in the
material to a single state variable $0 < r < 1$. This set of functions are able
to be efficiently evaluated by Horner's algorithm and serve as a close enough
approximation to the true non-linear conduction behavior for our
purposes.

In ReRAM, the overall resistance state as well as the transition behavior is
affected by a vast number of different possible configurations of ionic defects
in the material, giving rise to the observed stochastic behavior and history
dependence (Fig.~\ref{fig:correlated_cartoon}). Rather than
attempting to describe the ionic transport physically, we turn instead to
measurement data to directly provide the necessary statistical information. A
discrete multivariate stochastic process based on a Structural Vector
Autoregression (SVAR) model is fit to the data and used to generate latent
variables that guide the state evolution of simulated memory cells. As a cell is
exposed to voltage signals, new terms of the SVAR model are realized by a sum of
easily computable linear transformations of past states and pseudorandom
vectors.

As an overview, the experimental and simulation approach that will be elaborated in this
section can be shortly summarized as follows: 

\begin{enumerate}

\item A fabricated ReRAM cell is experimentally driven through a large number of
resistance cycles by applying a continuous periodic voltage signal while measuring the
resulting current.

\item A time series of feature vectors, $\x_n$, composed of resistance values and switching
threshold voltages, is extracted from each of the measured cycles.

\item A discrete stochastic process, $\x^*_n$, is constructed to enable generation of
simulated feature vectors that reproduce the measured distributions as
well as the long range correlation structure of $\x_n$.
 
\item An array of simulated cells are instantiated according to independent realizations of
$\x^*_n$ to represent cycle-to-cycle variations, together with a random scaling
vector $\bm{s}_m$ to represent device-to-device variations.

\item Two programming methods are exposed for each cell; one to apply voltages
and another to make realistic current readouts.
Applied voltages above the generated thresholds alter
the device state, following an empirical structure which
encodes the resistance transition behavior and allows access to a range of
resistance states. Each voltage driven resistance cycle triggers the generation
of new stochastic terms from $\x^*_n$, which govern the progression to future
states.

\end{enumerate}

\bigskip
\subsection{Data collection}

For the purposes of stochastic modelling, electrical measurement data is needed
that capture relevant information about the internal state of a memory cell and
its variation cycle-to-cycle (CtC) and device-to-device (DtD). However, ReRAM measurements
performed at operational speed typically make exclusive use of rectangular
voltage pulse sequences, which yield very little useful state information.
On the other hand, measurements applying continuously swept voltage signals while
sampling the resulting current are more suitable because much more information is
collected each cycle, such as switching threshold voltages, current-voltage
nonlinearity, resistance states, and transition behavior.

Conventionally, measurements employing voltage sweeps are
carried out using the source measure units (SMUs) of commercial
semiconductor parameter analyzers (SPAs).
However, SMUs make heavy use of averaging to measure noisy signals at
high resolution and thus sample too slowly to collect cycling data in a
meaningful quantity. Furthermore, because two-terminal switching devices are
prone to electrical instability and runaway transitions, voltage sweeping
measurements usually require integrated current limiting
transistors to avoid destruction or rapid degradation of the cell. This presents
a significant fabrication overhead and limits the materials available for
study. In light of these challenges, the input data for the present stochastic
model was acquired using a custom measurement technique, introduced in detail in
a recent publication \cite{hennen_current-limiting_2021}. The setup uses an
external current-limiting amplifier circuit to allow for collection of sweeping
measurements at over six orders of magnitude higher speeds than SMUs, while also
eliminating the cumbersome requirement of on-chip current
limiting.

\begin{figure}[h]
\begin{center}
  \includegraphics[width=10cm]{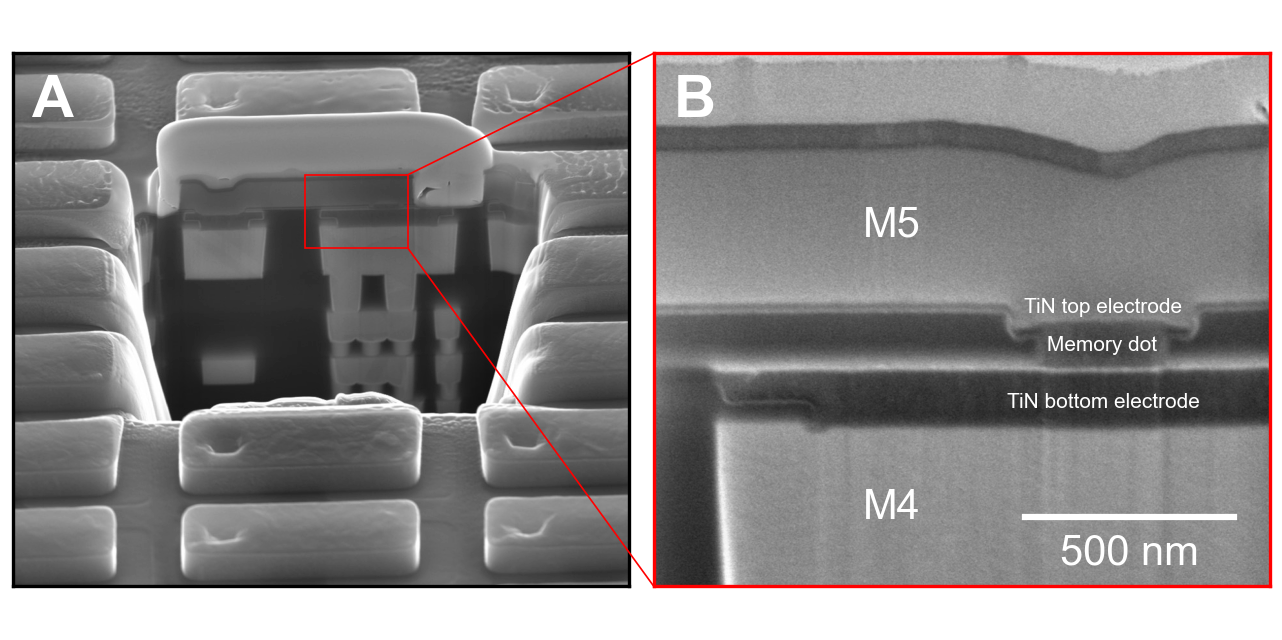}
\end{center}
\caption{Scanning electron micrographs of the ReRAM cell design used for
  electrical measurement. \textbf{(A)} shows a cross-section of the cell, and
\textbf{(B)} shows a zoom-in of the resistive memory between metalization layers
M4 and M5.
}\label{fig:SEM}
\end{figure}

The ReRAM cell used for measurement of cycling statistics (Fig.~\ref{fig:SEM}) was fabricated using a
combination of atomic layer deposition (ALD) and physical vapor deposition (PVD)
\cite{nail_understanding_2016}. The material stack from bottom to top was TiN /
ALD HfO$_2$ (10~nm) / PVD Ti (20~nm) / PVD TiN (100~nm). The device was
electrically isolated with contact pads leading directly to the top and bottom
device electrodes, using no access transistor or added series resistance. Using
a fixed 100~$\upmu$A current limit in the SET polarity,
the pristine cell was electroformed by application of
100~$\upmu$s duration triangular pulses with incrementally increasing amplitude
until a current jump was recorded near 3~V. For all subsequent cycling, a 1.5~V
amplitude 10~KHz triangular waveform was applied. The cell was first exercised
for $2.4 \times 10^6$ cycles before $10^6$ additional cycles were collected for
analysis. Current ($I$) and voltage ($U$) waveforms were simultaneously recorded
with 8-bit resolution and with a sample rate of 1,042 samples per cycle.

\begin{figure}[h]
\begin{center}
  \includegraphics[]{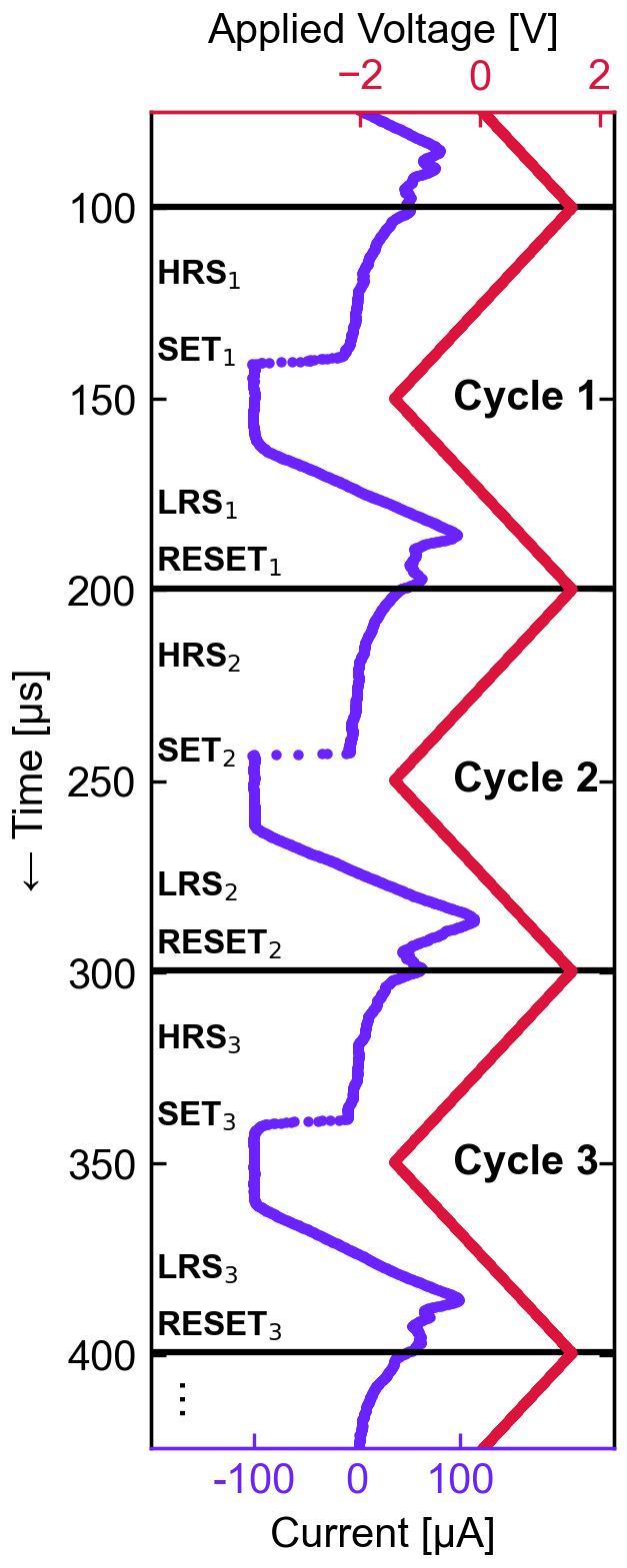}
\end{center}
\caption{
Measured time dependence of $I$ and $U$ waveforms resulting from the ReRAM
cycling experiment. The waveforms are divided into $10^6$ indexed cycles, the
first three of which are shown. From this dataset, the periodic temporal
sequence of the states and events of each cycle (HRS$_n$, SET$_n$, LRS$_n$,
RESET$_n$) is extracted and subject to statistical modelling.
}\label{fig:ivt_measured}
\end{figure}

The measured current array was smoothed with a moving average filter to improve
the quality of the raw data before further analysis. An adaptive rectangular
window size was used in order to preserve current steps in the signal, with the
maximum window size of 25 samples gradually reducing to a minimum
of 3 samples at the pre-detected locations of SET transitions of each cycle. After
smoothing, the contiguous $I$ and $U$ waveforms were split into indexible cycles
at most positive value of the periodic applied voltage
(see Fig.~\ref{fig:ivt_measured}).

\begin{figure}[h]
\begin{center}
\includegraphics[]{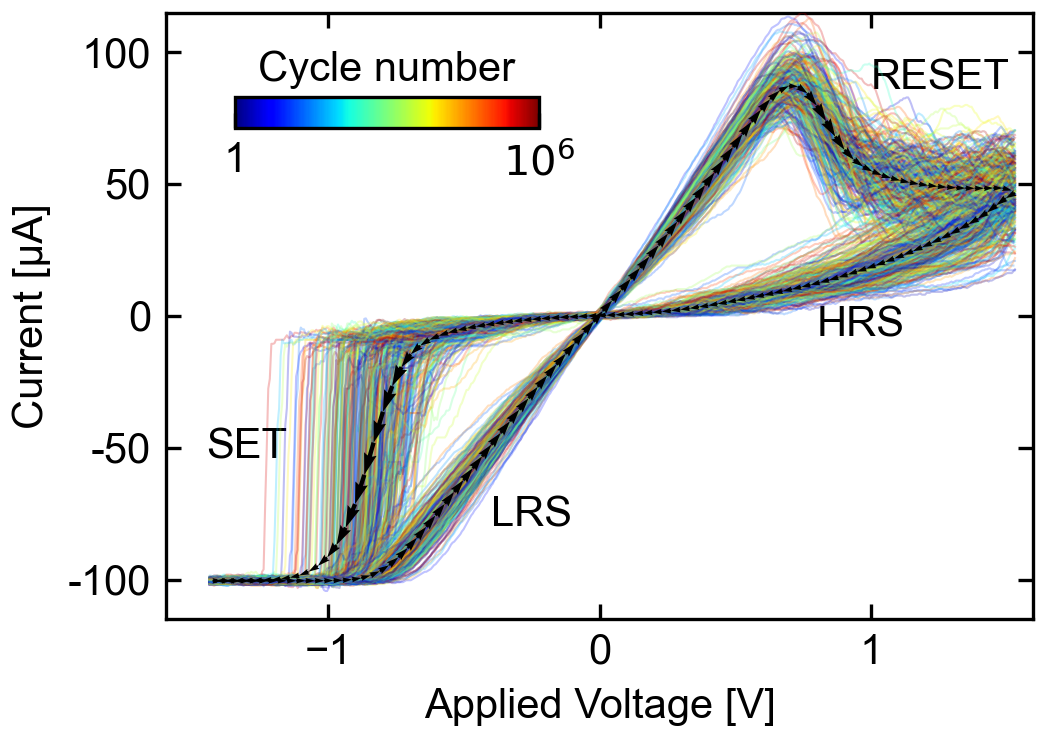}
\end{center}
\caption{A subset of the $10^6$ measured ($I,U$) cycles used as input to the
stochastic model.  The black arrowed path shows the average ($I,U$) curve and its temporal
direction. Different cycle indices are represented by
colored paths, which show significant statistical variation.
}\label{fig:ivloops_measured}

\end{figure}

Each cycle exhibits the following temporal sequence of states and events: a high
resistance state (HRS), a transition (SET) out of the HRS into the following low
resistance state (LRS), and finally another transition (RESET) into the next
HRS. Current vs. voltage ($I, U$) plots for a subset of the collected cycles are
shown in Fig.~\ref{fig:ivloops_measured}, which highlights the significant
stochastic CtC variations. The observed characteristics are typical
for ReRAM subjected to voltage controlled sweeps --- on average, there is relatively
higher voltage non-linearity in the HRS than in the LRS, and the SET transitions
are abrupt with respect to the applied voltage, while the RESET transitions
proceed relatively gradually over a voltage range of approximately 700~mV.

\bigskip
\subsection{Feature extraction}

The full $I, U$ cycling measurement just described consists of over 16~GB of
numerical data and would not be practical to model on a point-by-point basis.
Therefore, we aim to compress the dataset while retaining enough information
such that the full ($I,U$) characteristics can be approximately reconstructed from
the compressed representation. Accordingly, the full dataset is reduced
to a vector time series of
distinguishing features of each cycle. Four
scalar features were chosen for extraction: the value of the HRS, $R_H
[\Omega]$, the SET threshold voltage, $U_S [V]$, the value of the LRS, $R_L
[\Omega]$, and the RESET voltage, $U_R [V]$. We denote the series as

\begin{equation}
\x_n = 
\begin{bmatrix}
R_{H,n}\\ 
U_{S,n}\\ 
R_{L,n}\\ 
U_{R,n}
\end{bmatrix}
=
\begin{bmatrix}
R_{H}\\ 
U_{S}\\ 
R_{L}\\ 
U_{R}
\end{bmatrix}_n,
\end{equation}
where $n = \{1, 2, \ldots, 10^6\}$ is the set of cycle indices. The feature vector
elements, whose precise definition follows, are chronologically ordered from
top to bottom as they occur in the measurement dataset.

The SET voltage $U_S$, or the voltage where the cell resistance abruptly
decreases, is extracted from each cycle as the absolute value of the linearly
interpolated $U$ corresponding to the first level crossing of
$I=-50~\upmu\text{A}$. The RESET voltage $U_R$,
defined as the voltage where the reset process begins, is determined from the
$I$ datapoints by peak detection using simple comparison of neighboring samples.
Here, only the increasing section of the voltage sweep with $U>0$ is considered.
The voltage corresponding to the first encountered peak with prominence $\geq
5~\upmu$A
is taken as the RESET voltage. If no peak satisfies this criterion, the
peak with maximum prominence is taken instead.

The device current for any static
state is approximated in our model as a polynomial function of
the applied voltage. The values of $R_H$ and $R_L$ are likewise extracted
from least squares polynomial fits to appropriate
subsets of the measured $(I,U)$ data of each cycle. The HRS
is fit with a 5th degree polynomial on the decreasing $U$ sweep in the variable
range $U_S+0.1~\text{V}~\leq~U~\leq~1.5~\text{V}$ and
$-25~\upmu\text{A}~\leq~I~\leq~80~\upmu\text{A}$, and the LRS is fit with a 3rd
degree polynomial on the increasing part of the $V$ sweep in the range
$-0.7~\text{V}~\leq~U~\leq~U_R-0.05~\text{V}$ and
$-80~\upmu\text{A}~\leq~I~\leq~120~\upmu\text{A}$. The fits are constrained such
that the 0th order coefficient equals 0~A, and the 1st order coefficient is
$\geq$1~nA/V. The values of $R_H$ and $R_L$ are then defined as the static
resistance of the respective polynomials at a fixed voltage $U_0=200$~mV.

\begin{figure}[h]
\begin{center}
  \includegraphics[]{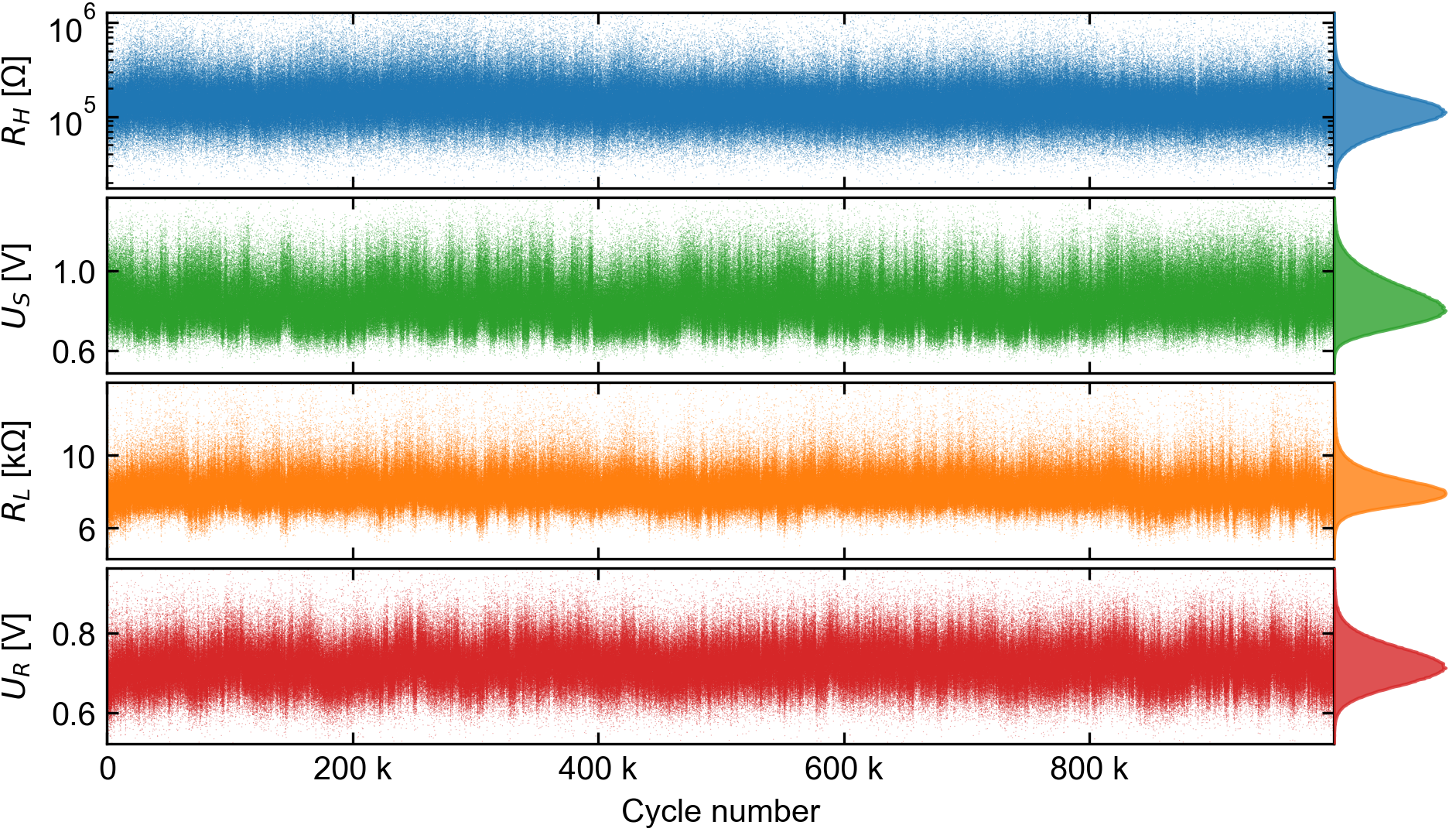}
\end{center}
\caption{A view of the feature vector time series extracted from each of $10^6$ measured ($I,U$) cycles. Each
  feature, which represents either a resistance state or a switching voltage, has its marginal histogram shown on the right.
  }
\label{fig:1M_parameters}
\end{figure}

An overview of the result of this feature extraction is given in
Fig.~\ref{fig:1M_parameters}. The $10^6$ cycles proceeded
without significant long-term drift from the overall mean value,
\begin{equation}
\label{eq:xbar}
\bar{\x}_n = 
\begin{bmatrix}
166.5~\text{k}\Omega\\ 
0.85~\text{V}\\ 
8.2~\text{k}\Omega\\ 
0.72~\text{V}
\end{bmatrix},
\end{equation}
but with significant variations in each feature between cycles. A prominent
characteristic of this data is that it is strongly correlated over long
cycle ranges, as quantified in
Fig.~\ref{fig:autocorrelation}. The asymmetric marginal distributions for each
of the features were very well resolved due to the large
number of samples, and they did not accurately converge to any analytical
probability density function (PDF) in common use, including the normal and
log-normal.

\bigskip
\subsection{Stochastic Modelling}

\begin{figure}[h]
\begin{center}


\def\dx{8.5mm}
\def\dy{8.5mm}

\tikzstyle{constant}          = [rectangle, minimum width=\dx, minimum height=\dy, inner sep=0mm, text centered, text width=\dx, draw=black, fill=white]
\tikzstyle{constant1d}        = [rectangle, minimum width=\dx, minimum height=\dy, inner sep=0mm, text centered, text width=\dx, draw=black, fill=white, thick]
\tikzstyle{constant2d}        = [rectangle, minimum width=\dx, minimum height=\dy, inner sep=0mm, text centered, text width=\dx, draw=black, fill=white, double distance=.3mm]
\tikzstyle{constant3d}        = [rectangle, minimum width=\dx, minimum height=\dy, inner sep=0mm, text centered, text width=\dx, draw=black, fill=white, double distance=.3mm, thick] 
\tikzstyle{latent}            = [circle, minimum width=1cm, minimum height=1cm, inner sep=0mm, text centered, text width=1cm, draw=black, fill=white]
\tikzstyle{latent1d}          = [circle, minimum width=1cm, minimum height=1cm, inner sep=0mm, text centered, text width=1cm, draw=black, fill=white, thick]
\tikzstyle{deterministic}     = [circle, dashed, minimum width=1cm, minimum height=1cm, inner sep=0mm, text centered, text width=1cm, draw=black, fill=white]
\tikzstyle{deterministic1d}   = [circle, dashed, minimum width=1cm, minimum height=1cm, inner sep=0mm, text centered, text width=1cm, draw=black, fill=white, thick]
\tikzstyle{observed}          = [circle, minimum width=1cm, minimum height=1cm, inner sep=0mm, text centered, text width=1cm, draw=black, fill=gray!50]
\tikzstyle{function}          = [rectangle, rounded corners, minimum width=.8cm, minimum height=.8cm, inner sep=0mm, text centered, text width=.8cm, draw=black, fill=white]
\tikzstyle{plate}             = [rectangle, rounded corners, draw, thick]

\tikzstyle{arrow} = [very thick,->,>=stealth]

\usetikzlibrary{decorations.pathreplacing}
\begin{tikzpicture}[node distance=4mm]


\node (a) [constant] {$a$};
\node (Sigma) [constant2d, right=of a] {$\bm{\Sigma}$};
\node (gamma) [constant2d, right=of Sigma] {$\bm{\gamma}$};
\node (A) [constant2d, right=of gamma] {$\bm{A}$};
\node (B) [constant2d, right=of A] {$\bm{B}$};
\node (Ci) [constant3d, right=of B] {$\bm{C}_i$};
\node (Umax) [constant, right=of Ci, xshift=0cm] {$U_{\text{max}}$};
\node (p) [constant1d, right=of Umax] {$\bm{p}$};
\node (Uread) [constant, right=of p] {$U_{\text{read}}$};
\node (Df) [constant, right=of Uread] {$\Delta f$};
\node (Imin) [constant, right=of Df] {$I_{\text{min}}$};
\node (Imax) [constant, right=of Imin] {$I_{\text{max}}$};
\node (nbits) [constant, right=of Imax] {$n_{\text{bits}}$};

\node (shat) [latent1d, below=of a, yshift=-1cm, xshift=0.6cm] {$\widehat{\bm{s}}_m$};
\node (Gamma1) [function, below=of shat, yshift=-0.5cm] {$\bm{\Gamma}\inv$};
\node (s) [deterministic1d, right=of Gamma1] {$\bm{s}_m$};
\node (eps) [latent1d, right=of s, yshift=1.5 cm +.5 cm] {$\bm{\epsilon}_n$};
\node (xhat) [deterministic1d, right=of eps] {$\widehat{\bm{x}}^*_n$};
\node (Gamma2) [function, right=of xhat] {$\bm{\Gamma}\inv$};
\node (x) [deterministic1d, right=of Gamma2] {$\bm{x}^*_n$};
\node (y) [deterministic1d, below=of x] {$\bm{y}^*_{m,n}$};
\node (r) [deterministic, right=of x] {$r$};
\node (U) [constant, below=of r, yshift=-0.15cm] {$U_a$};
\node (I) [deterministic, right=of r] {$I_{\text{read}}$};
\node (sigma) [deterministic, right=of I] {$\sigma_I$};
\node (IADC) [observed, right=of sigma, yshift=-1.25cm] {$I_{\text{ADC}}$};

\draw [plate] (-.25,-1) rectangle ($(IADC.east) + (1.35,-2.3)$) node[pos=.9, xshift=-.5cm] {\Large$\forall m \in \{1, \ldots, M\}$};
\draw [plate] ($(eps.west)!.5!(s.east) + (0, 1.55 + .25)$) rectangle ($(IADC.east) + (.5, -1.4)$) node[pos=.86, xshift=-3mm] {\Large$\forall n \in \{1, \ldots, N\}$};

\draw [arrow] (a) to[out=-90, in=110] node[] {} (shat);
\draw [arrow] (Sigma) to[out=-90, in=70] node[] {} (shat);
\draw [arrow] (gamma) to[out=-100,in=60] node[] {} (Gamma1);
\draw [arrow] (gamma) to[out=-80, in=120] node[] {} (Gamma2);
\draw [arrow] (A) to[out=-70, in=120] node[] {} (xhat);
\draw [arrow] (B) to[out=-90, in=90] node[] {} (xhat);
\draw [arrow] (Ci) to[out=-120, in=60] node[] {} (xhat);
\draw [arrow] (Umax) to[out=-60,in=120] node[] {} (r);
\draw [arrow] (p) to[out=-70,in=100] node[] {} (r);
\draw [arrow] (p) to[out=-65,in=120] node[] {} (I);
\draw [arrow] (Uread) to[out=-70,in=100] node[] {} (I);
\draw [arrow] (Uread) to[out=-65,in=120] node[] {} (sigma);
\draw [arrow] (Df) to[out=-70,in=100] node[] {} (sigma);
\draw [arrow] (Imin) to[out=-75,in=112.5] node[] {} (IADC);
\draw [arrow] (Imax) to[out=-90, in=90] node[] {} (IADC);
\draw [arrow] (nbits) to[out=-105,in=67.5] node[] {} (IADC);

\draw [arrow] (shat) -- node[] {} (Gamma1);
\draw [arrow] (Gamma1) -- node[] {} (s);
\draw [arrow] (s) to[out=0, in=180] node[] {} (y);
\draw [arrow] (eps) -- node[] {} (xhat);
\draw [arrow, dotted] (xhat) to [out=-80, in=-130, looseness=3]  node [black,midway,below=-1mm, anchor=north]{\footnotesize$n \mapsto \{n+1, \ldots, n+p\}$} (xhat);

\draw [arrow] (xhat) -- node[] {} (Gamma2);
\draw [arrow] (Gamma2) -- node[] {} (x);
\draw [arrow] (x) -- node[] {} (y);
\draw [arrow] (y) -- node[] {} (r);
\draw [arrow] (U) -- node[] {} (r);
\draw [arrow] (r) -- node[] {} (I);
\draw [arrow] (I) -- node[] {} (sigma);
\draw [arrow] (I) to[out=-50, in=157.5] node[] {} (IADC);
\draw [arrow] (sigma) to[out=-45, in=135] node[] {} (IADC);

\node (table1) [below=of a, yshift=-5.2cm, xshift=-0.25, anchor=north west]{
\begin{footnotesize}
\begin{tabular}{l l l l}
$a$ & Scale factor for $\bm{\Sigma}$                      & $\widehat{\bm{s}}_m$ & $\N(\bm{0}, a\Sigma)$ sample\\
$\bm{\Sigma}$ & Device covariance                    & $\G\inv$ & Inverse normalizing map \\
$\bm{\gamma}$ & Coefficients for $\G\inv$   & $\bm{s}_m$ & Device scale vector \\
$\bm{A}$ & SVAR contemp. parameters                  & $\bm{\epsilon}_n$ & White noise vector \\
$\bm{B}$ & SVAR noise amplitudes                     & $\xhat^*_n$  & SVAR process\\
$\bm{C}_i$ & SVAR lag parameters                     & $\x^*_n$ & Median cycling process \\
$U_\text{max}$ & Max. voltage applied                & $\bm{y}^*_{m,n}$ & Device cycling process\\
$\bm{p}$ & HHRS, LLRS coefficients                  & $r$ & Device state variable \\
$U_\text{read}$ & Readout voltage                    & $U_a$ &  Applied voltage pulse(s) \\
$\Delta f$ & Readout bandwidth                       & $I_{\text{read}}$ & Readout current (noiseless)\\
$I_\text{min}$ & Min. current of ADC                   & $\sigma_I$ & Readout noise amplitude \\
$I_\text{max}$ & Max. current of ADC                   & $I_\text{ADC}$ & ADC readout \\
$n_\text{bits}$ & Number of ADC bits                 & &
\end{tabular}
\end{footnotesize}
};

\matrix (m) [matrix of nodes, right=of table1, xshift=-.5cm, yshift=.25cm, column 1/.style={anchor=center}, column 2/.style={anchor=west}, nodes={rectangle, minimum height=7mm}] {
      \node[constant, scale=.5] {}; & \footnotesize Constant node \\
      \node[latent, scale=.5] {}; & \footnotesize Stochastic node \\
      \node[deterministic, scale=.5, densely dashed] {}; & \footnotesize Deterministic node \\
      \node[observed, scale=.5] {}; & \footnotesize Observed node \\
      \node[function, scale=.5, rounded corners=.7mm] {}; & \footnotesize Function \\
      \node[plate, minimum width=.8cm, minimum height=.8cm] {}; & \footnotesize Plate \\
      };

\matrix (m2) [matrix of nodes, right=of m, xshift=-.25cm, column 1/.style={anchor=center}, column 2/.style={anchor=west}, nodes={rectangle, minimum height=7mm}] {
      \draw (0,0) -- (1,0); & \footnotesize 0D \\
      \draw[thick] (0,0) -- (1,0); & \footnotesize 1D \\
      \draw[double distance=.3mm] (0,0) -- (1,0); & \footnotesize 2D \\
      \draw[double distance=.3mm, thick] (0,0) -- (1,0); & \footnotesize 3D \\
      };

\end{tikzpicture}
\caption{
  Graphical model depicting the relationships between all parameters and latent
variables involved in the stochastic synapse model. Plate notation is used to
represent $N$ switching cycles of $M$ devices, each yielding an observed readout current.
The dotted recurrent arrow denotes a connection to each of the $p$ following frames,
as needed by the history dependent stochastic process.
}\label{fig:graphical_model}
\end{center}
\end{figure}
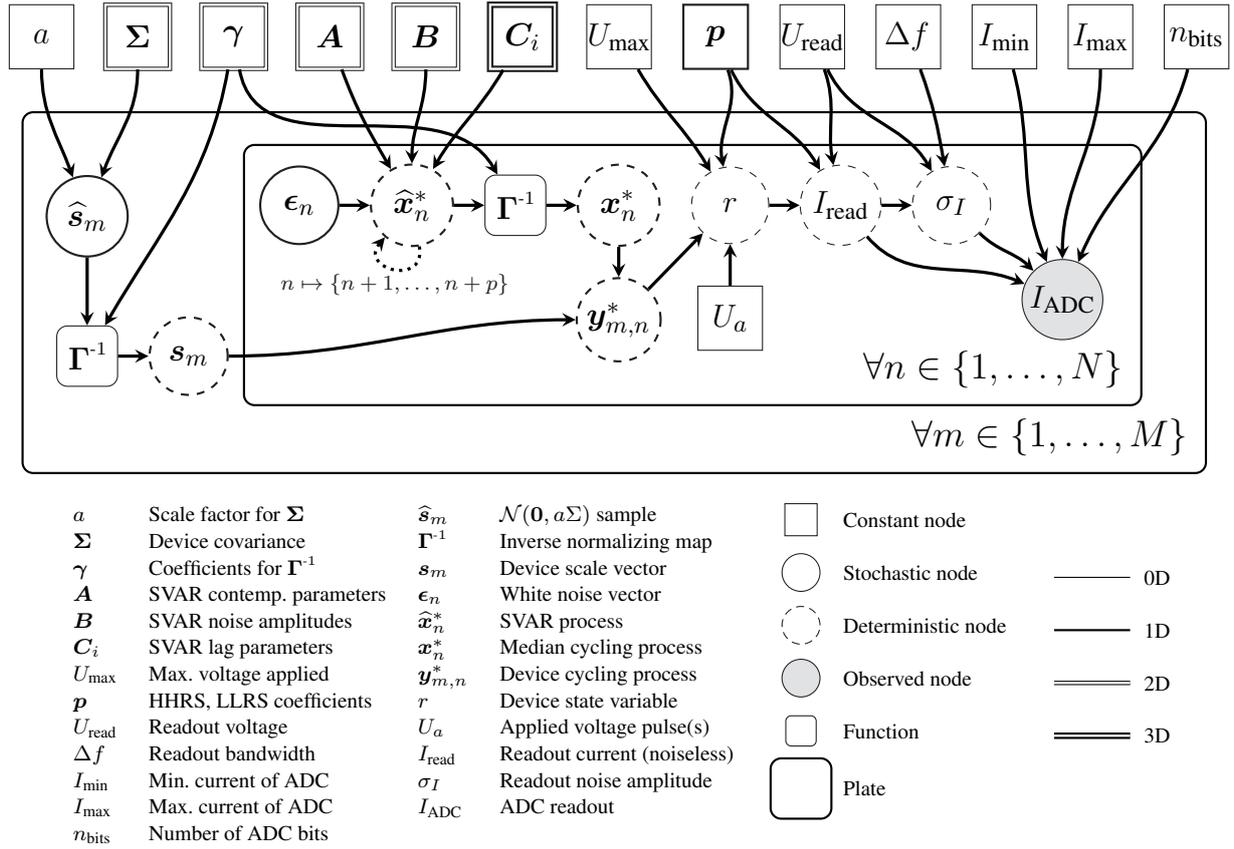

This section will introduce the statistical methods used to model the internal
states of an array of synaptic ReRAM devices, including CtC and DtD variability
effects. The handling of voltages applied to the cells as well as the simulation
of realistic readouts of the resistance states will also be established. To help
\comment{dis}orient the reader, the overall structure of the generative model that will be
described is provided in advance in Fig.~\ref{fig:graphical_model}.

\bigskip
\subsubsection{Cycle-to-cycle (CtC) variations}

In seeking to represent the input time series $\x_n$ with a stochastic
process, the main goals are to recreate the marginal
distributions as well as the correlation structure of its vector components. To
achieve the first goal with high generality, we use an approach based on
transformation of the measured densities to and from the standard normal
distribution $\N(0,1)$. This way, a single process can be used
to achieve any set of marginals presented by the input data, with the relatively
unrestrictive requirement that this base process generates normal marginals.
Notationally, we define and apply an invertible, smooth mapping $\G : \R^4
\rightarrow \R^4$ that normalizes the marginal distributions of the vector
components,

\begin{equation}\label{eq:gamma}
\x_n  
=
\begin{bmatrix}
R_H\\ 
U_S\\ 
R_L\\ 
U_R
\end{bmatrix}_n
\xrightarrow{\G}
\begin{bmatrix}
\widehat{R}_H\\ 
\widehat{U}_S\\ 
\widehat{R}_L\\ 
\widehat{U}_R
\end{bmatrix}_n
=
\xhat_n,
\end{equation}
where a hatted variable signifies that it is distributed as $\N(0,1)$.
We then construct a base process $\xhat^*_n$ whose marginals are normal,
and finally transform its output back to the original data distributions via the
inverse map $\G\inv$. The overall process $\x^*_n$ is thus defined,

\begin{equation}\label{eq:gammainverse}
\xhat^*_n  
=
\begin{bmatrix}
\widehat{R}_H^*\\ 
\widehat{U}_S^*\\ 
\widehat{R}_L^*\\ 
\widehat{U}_R^*
\end{bmatrix}_n
\xrightarrow{\G\inv}
\begin{bmatrix}
R_H^*\\ 
U_S^*\\ 
R_L^*\\ 
U_R^*
\end{bmatrix}_n
=
\x^*_n,
\end{equation}
where a star indicates a generated random variable to distinguish from variables
originating from measurement data.

\begin{figure}[h]
\begin{center}
  \includegraphics[]{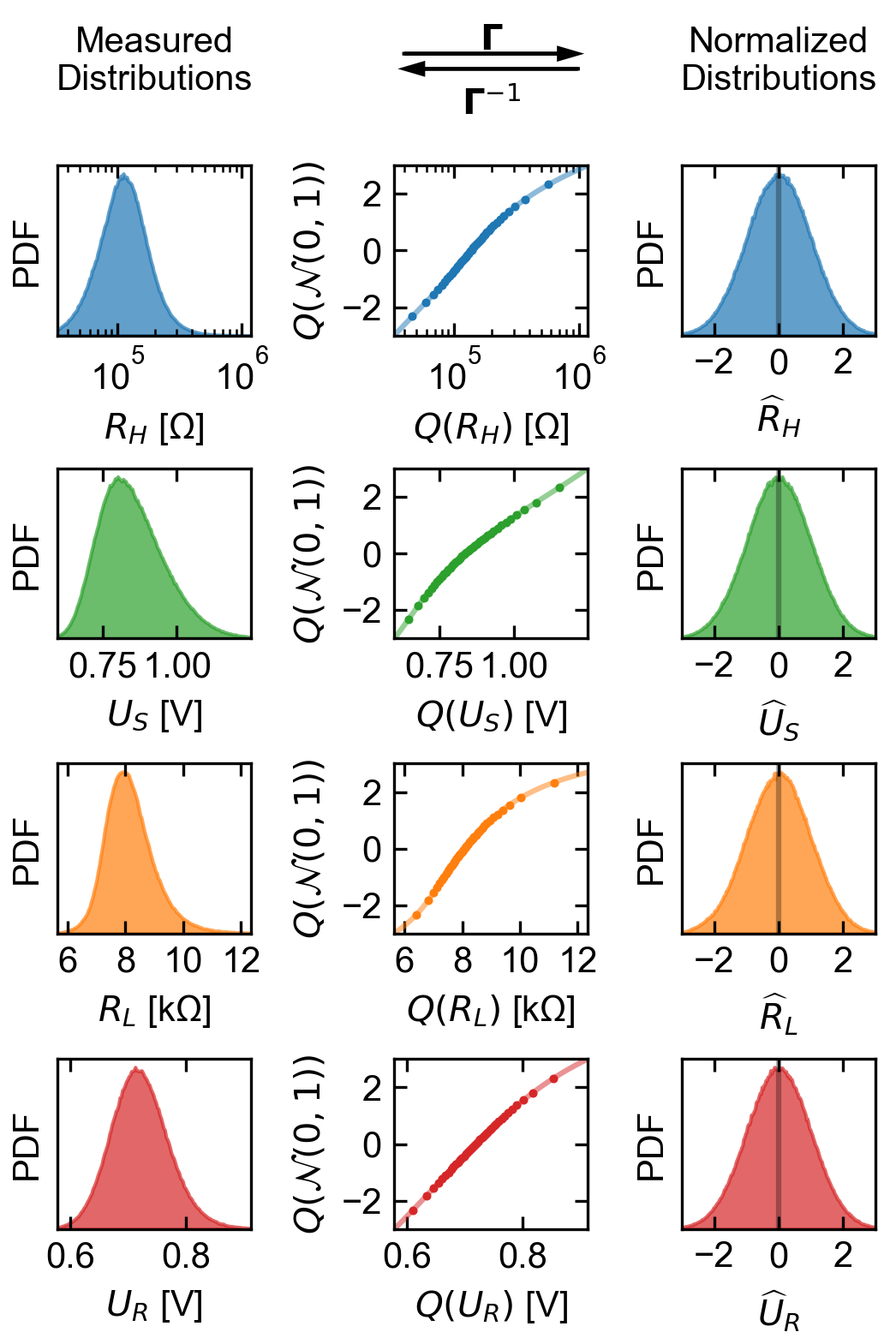}
\end{center}
\caption{Visualization of the invertible normalizing transformation $\G$ that is
applied to the measured feature vectors before fitting with a base stochastic
process. The left column shows the marginal PDFs of the vector time series
$\x_n$ extracted from measurement. The center column shows the input and output
quantile-quantile plots with the fitted log-polynomial function used to
transform the distributions (here, $Q$ denotes the quantile function of its
argument). The right column is the result of applying $\G$ to the input data,
producing $\xhat_n$ whose elements are normally distributed.
}\label{fig:normalizing_map}
\end{figure}

This type of density transformation procedure is a widely used technique for
working with arbitrary distributions, which finds
application in a variety of fields and can be constructed in many different
ways \cite{cario_autoregressive_1996, rezende_variational_2015}. While the
transformation is trivially constructed in the case where the target quantile
function and its inverse are each analytically defined, we do not make this
assumption in the present scenario. A simple numerical method in this case is a
so-called quantile transform, where the input and output quantile functions are
each discretely sampled and the transformation is defined through a direct map
between bins or through interpolation. The main requirement for $\G$ in our
model, however, is that its inverse (Eq.~\ref{eq:gammainverse}) is easy to
evaluate without causing cache misses due to memory access, thus it is
preferable to avoid referencing and interpolation of large look-up tables. The
forward transformation (Eq.~\ref{eq:gamma}), on the other hand, only needs to be
computed once for model fitting and is not used for the generating process. We
therefore define $\G\inv$ as essentially a quantile transform, operating on each
feature independently, that is evaluated from a fit of the quantiles to a
specific analytic function. Namely,
\begin{equation}
\bm{\G}\inv(\widehat\x_n)
=
\exp
\begin{bmatrix*}[l]
\gamma_1(\widehat{R}_{H,n})\\ 
\gamma_2(\widehat{U}_{S,n})\\ 
\gamma_3(\widehat{R}_{L,n})\\ 
\gamma_4(\widehat{U}_{R,n})
\end{bmatrix*}
=
\x_n,
\end{equation}
where $\gamma_1$-$\gamma_4$ are each 5th degree polynomials, and the exponential
function is applied element-wise. The coefficients of the polynomials are fit to
standard normal quantiles vs. those of the respective (log) features, sampled
at 500 equally spaced values between 0.01 and 0.99. The fitted polynomials are
checked for monotonicity within four standard deviations above and below zero, and
the forward transformation, 
\begin{equation}
\bm{\G}(\x_n)
=
\begin{bmatrix*}[l]
\gamma_1\inv(\log{R_{H,n}})\\ 
\gamma_2\inv(\log{U_{S,n}})\\ 
\gamma_3\inv(\log{R_{L,n}})\\ 
\gamma_4\inv(\log{U_{R,n}})
\end{bmatrix*}
= \xhat_n
,
\end{equation}
is computed using numerical inverse of the $\gamma$ polynomials. A visualization
of the function $\G$ as well as the marginal histograms corresponding to input
series $\x_n$ and output series $\xhat_n$, are shown in Fig.~\ref{fig:normalizing_map}.

Now that we have transformed the input measurement data into a normalized vector
time series $\xhat_n$, a suitable stochastic process will be chosen
for fitting.
This process should serve as a useful approximation to the true physical
mechanisms that generated the data, capturing the long-range correlation
structure of the observed features. Time series analysis is broadly used across
scientific and engineering domains, but despite its applicability to the rich
statistical behavior displayed by resistive switching devices, device
models have not yet widely employed dependent
stochastic processes. Many models and analyses assume for convenience that
features are independently and identically distributed
according to a normal or lognormal PDF \cite{li_device_2017,
chen_utilizing_2015}. However, there is not a strong theoretical basis for
this assumption in a highly nonlinear and path-dependent system based on
continuous evolution of conducting filaments. Dependent stochastic processes, on
the other hand, more appropriately allow for a description of the dependence of
future states on past states.

Simple models in the category of Markov chains have been considered as
generating processes for memory cells. A rudimentary example is a 1-dimensional
random walk process, where each future state is computed as a random additive
perturbation on the previous state
\cite{bengel_variability-aware_2020}. While random walk
represents a reasonable short range approximation, it has
the well known property that the expected absolute distance between the initial
value and the $N$th value is proportional to $\sqrt{N}$ for large
$N$, causing the
process to eventually drift to unphysical values without the use of artificial
constraints.

Autoregressive (AR) models are simple univariate processes sharing some
characteristics of random walk, but based additionally on a deterministic linear
dependence on past observations. Each new term of an AR($p$) (AR
of order $p$) model is computed by linear combinations of $p$ previous (lagged)
values together with a noise term, producing processes that are wide-sense
stationary and mean-reverting within suitable parameter ranges
\cite{hamilton_time_1994, lutkepohl_new_2005}.
The few times they have appeared in the literature, low order models like AR(1)
and AR(2) were used to describe state variables independently (e.g. a
sequence of high and/or low resistance states) \cite{Fantini_intrinsic_2015,
roldan_time_2019}. Here we pursue a more
comprehensive statistical description of the interrelations between the
different variables contained in the vectors $\xhat_n$ which takes into account
long range correlations $p \gg 1$. This is enabled by the use of a VAR($p$)
model (vector AR of order $p$), which is the multivariate counterpart of the AR
model applicable to discrete vector time series \cite{hamilton_time_1994,
lutkepohl_new_2005}.

\begin{figure}[h]
\begin{center}
\def \Aaa {-1.000}
\def \Aab {-0.000}
\def \Aac {-0.000}
\def \Aad {-0.000}
\def \Aba {0.111}
\def \Abb {-1.000}
\def \Abc {-0.000}
\def \Abd {-0.000}
\def \Aca {0.023}
\def \Acb {-0.139}
\def \Acc {-1.000}
\def \Acd {-0.000}
\def \Ada {-0.008}
\def \Adb {0.070}
\def \Adc {0.180}
\def \Add {-1.000}
\def \Add {1.000}
\def \Baa {0.984}
\def \Bab {0.000}
\def \Bac {0.000}
\def \Bad {0.000}
\def \Bba {0.000}
\def \Bbb {0.945}
\def \Bbc {0.000}
\def \Bbd {0.000}
\def \Bca {0.000}
\def \Bcb {0.000}
\def \Bcc {0.908}
\def \Bcd {0.000}
\def \Bda {0.000}
\def \Bdb {0.000}
\def \Bdc {0.000}
\def \Bdd {0.921}
\def \Caa {0.043}
\def \Cab {0.021}
\def \Cac {0.037}
\def \Cad {-0.002}
\def \Cba {0.015}
\def \Cbb {0.057}
\def \Cbc {0.028}
\def \Cbd {-0.011}
\def \Cca {0.010}
\def \Ccb {-0.000}
\def \Ccc {0.153}
\def \Ccd {0.010}
\def \Cda {0.001}
\def \Cdb {0.002}
\def \Cdc {0.023}
\def \Cdd {0.085}
\def \lwAaa {1.500mm}
\def \lwAab {0.300mm}
\def \lwAac {0.300mm}
\def \lwAad {0.300mm}
\def \lwAba {1.042mm}
\def \lwAbb {1.500mm}
\def \lwAbc {0.300mm}
\def \lwAbd {0.300mm}
\def \lwAca {0.454mm}
\def \lwAcb {1.226mm}
\def \lwAcc {1.500mm}
\def \lwAcd {0.300mm}
\def \lwAda {0.351mm}
\def \lwAdb {0.769mm}
\def \lwAdc {1.497mm}
\def \lwAdd {1.500mm}
\def \lwCaa {0.584mm}
\def \lwCab {0.439mm}
\def \lwCac {0.547mm}
\def \lwCad {0.316mm}
\def \lwCba {0.399mm}
\def \lwCbb {0.678mm}
\def \lwCbc {0.487mm}
\def \lwCbd {0.372mm}
\def \lwCca {0.366mm}
\def \lwCcb {0.302mm}
\def \lwCcc {1.318mm}
\def \lwCcd {0.366mm}
\def \lwCda {0.307mm}
\def \lwCdb {0.311mm}
\def \lwCdc {0.453mm}
\def \lwCdd {0.865mm}

\definecolor{tabblue}{HTML}{1F77B4}
\definecolor{taborange}{HTML}{FF7F0E}
\definecolor{tabgreen}{HTML}{2CA02C}
\definecolor{tabred}{HTML}{D62728}



\tikzstyle{HRS}   = [rectangle, minimum width=2cm, minimum height=1cm, inner sep=0mm, text centered, text width=1cm, draw=black, fill=tabblue!50]
\tikzstyle{US}    = [rectangle, minimum width=2cm, minimum height=1cm, inner sep=0mm, text centered, text width=1cm, draw=black, fill=tabgreen!50]
\tikzstyle{LRS}   = [rectangle, minimum width=2cm, minimum height=1cm, inner sep=0mm, text centered, text width=1cm, draw=black, fill=taborange!50]
\tikzstyle{UR}    = [rectangle, minimum width=2cm, minimum height=1cm, inner sep=0mm, text centered, text width=1cm, draw=black, fill=tabred!50]
\tikzstyle{noise} = [rectangle, minimum width=1cm, minimum height=1cm, inner sep=0mm, text centered, text width=1cm, draw=black, fill=black!10]

\tikzstyle{arrow} = [very thick,->,>=stealth]
\tikzstyle{redarrow} = [arrow, color=tabred]
\tikzstyle{orangearrow} = [arrow, color=taborange]
\tikzstyle{bluearrow} = [arrow, color=tabblue]
\tikzstyle{greenarrow} = [arrow, color=tabgreen]

\usetikzlibrary{decorations.pathreplacing}
\begin{tikzpicture}[node distance=0.5cm and 0.2cm]


\def\dx{3mm}
\def\dy{2mm}

\node (HRS0) [HRS] {$\widehat{R}^*_{H,n-1}$};
\node (US0)  [US, below right=of HRS0] {$\widehat{U}^*_{S,n-1}$};
\node (LRS0) [LRS, below right=of US0] {$\widehat{R}^*_{L,n-1}$};
\node (UR0)  [UR, below right=of LRS0] {$\widehat{U}^*_{R,n-1}$};
\node (HRS1) [HRS, right=of HRS0, xshift=7cm] {$\widehat{R}^*_{H,n}$};
\node (US1)  [US, below right=of HRS1] {$\widehat{U}^*_{S,n}$};
\node (LRS1) [LRS, below right=of US1] {$\widehat{R}^*_{L,n}$};
\node (UR1)  [UR, below right=of LRS1] {$\widehat{U}^*_{R,n}$};
\node (eps0) [noise, above=of HRS1, yshift=0.5cm] {$\epsilon_{1,n}$};
\node (eps1) [noise, above=of US1, yshift=2cm] {$\epsilon_{2,n}$};
\node (eps2) [noise, above=of LRS1, yshift=3.5cm] {$\epsilon_{3,n}$};
\node (eps3) [noise, above=of UR1, yshift=5cm] {$\epsilon_{4,n}$};


\draw [bluearrow, line width=\lwCad] ([yshift=3*\dy/2]   UR0.east) -| node[xshift=-0.6cm, yshift=0, anchor=south]{\Cad}([xshift=0]HRS1.south);
\draw [bluearrow, line width=\lwCac] ([yshift=\dy]      LRS0.east) -| node[xshift=-0.6cm, yshift=0, anchor=south]{\Cac}([xshift=-\dx]HRS1.south);
\draw [bluearrow, line width=\lwCab] ([yshift=\dy/2]     US0.east) -| node[xshift=-0.6cm, yshift=0, anchor=south]{\Cab}([xshift=-2*\dx]HRS1.south);
\draw [bluearrow, line width=\lwCaa] (HRS0.east) -- node[pos=.92, xshift=-.1cm, yshift=0, anchor=north]{\Caa} (HRS1);
\draw [bluearrow,  line width=1mm] (eps0) node[yshift=-1cm, anchor=east]{\Baa} -- (HRS1);

\draw [greenarrow, line width=\lwAba] ([yshift=-\dy]    HRS1.east) -| node[xshift=-0.6cm, yshift=-0.2cm, anchor=north]{\Aba}([xshift=-\dx]US1.north);
\draw [greenarrow, line width=\lwCbd] ([yshift=\dy/2]  UR0.east) -| node[xshift=-0.6cm, yshift=0cm, anchor=south]{\Cbd}([xshift=0]US1.south);
\draw [greenarrow, line width=\lwCbc] ([yshift=0     ] LRS0.east) -| node[xshift=-0.6cm, yshift=0cm, anchor=south]{\Cbc}([xshift=-\dx]US1.south);
\draw [greenarrow, line width=\lwCbb] ([yshift=-\dy/2]   US0.east) -- node[pos=.93, anchor=north]{\Cbb}([yshift=-\dy/2]US1.west);
\draw [greenarrow, line width=1mm]    (eps1) node[yshift=-1cm, anchor=east]{\Bbb} --   (US1);

\draw [orangearrow, line width=\lwAcb] ([yshift=-\dy/2] US1.east) -| node[xshift=-0.7cm, yshift=-.35cm, anchor=north]{\Acb}([xshift=-2*\dx]LRS1.north);
\draw [orangearrow, line width=\lwAca] ([yshift=0    ]HRS1.east) -| node[xshift=-0.5cm,         anchor=north]{\Aca} ([xshift=-\dx]LRS1.north);
\draw [orangearrow, line width=\lwCcd] ([yshift=-\dy/2] UR0.east) -| node[xshift=-0.6cm,         anchor=south]{\Ccd}([xshift=-0]LRS1.south);
\draw [orangearrow, line width=\lwCcc]([yshift=-\dy]  LRS0.east) --  node[pos=.92, yshift=-.1cm,  anchor=north]{\Ccc}([yshift=-\dy]LRS1.west);
\draw [orangearrow, line width=1mm] (eps2) node[yshift=-1cm, anchor=east]{\Bcc} -- (LRS1);

\draw [redarrow, line width=\lwAdc] ([yshift=0   ]   LRS1.east)  -| node[xshift=-0.6cm, yshift=-.5cm, anchor=north]{\Adc} ([xshift=-3*\dx]UR1.north);
\draw [redarrow, line width=\lwAdb] ([yshift=\dy/2]  US1.east)  -| node[xshift=-0.6cm, anchor=north]{\Adb} ([xshift=-2*\dx]   UR1.north);
\draw [redarrow, line width=\lwAda] ([yshift=\dy]   HRS1.east)  -| node[xshift=-0.6cm, anchor=north]{\Ada} ([xshift=-1*\dx] UR1.north);
\draw [redarrow, line width=\lwCdd] ([yshift=-3*\dy/2] UR0.east)  -- node[pos=.92, xshift=-0, yshift=.05cm, anchor=south]{\Cdd} ([yshift=-3*\dy/2] UR1.west);
\draw [redarrow, line width=1mm]    (eps3) node[yshift=-1cm, anchor=east]{\Bdd} -- (UR1);


\draw [decorate, thick, decoration={brace, amplitude=.3cm,mirror,raise=4pt}] (-1,-5.1) -- (7.7,-5.1) node [black,midway,below=.5cm]{\large $\widehat{\bm{x}}^*_{n-1}$};
\draw [decorate, thick, decoration={brace, amplitude=.3cm,mirror,raise=4pt}] (8.2,-5.1) -- (16.9,-5.1) node [black,midway,below=.5cm]{\large $\widehat{\bm{x}}^*_{n}$};



\end{tikzpicture}
\caption{A weighted graph displaying the causal structure of the utilized SVAR($p$)
process, showing the nearest temporal contributions to realizations of the
random vector $\xhat^*_n$. Arrow weights show the model
parameters contained in $\bm{A}$, $\bm{B}$ and the upper triangular part of
$\bm{C}_1$ when fit with $p=100$. The actual SVAR($p$) model uses many more
connections than shown ($16p + 10$), so that each variable is impacted by all past values of
all other variables within cycle
range $p$.
}\label{fig:structural_flowchart}
\end{center}
\end{figure}
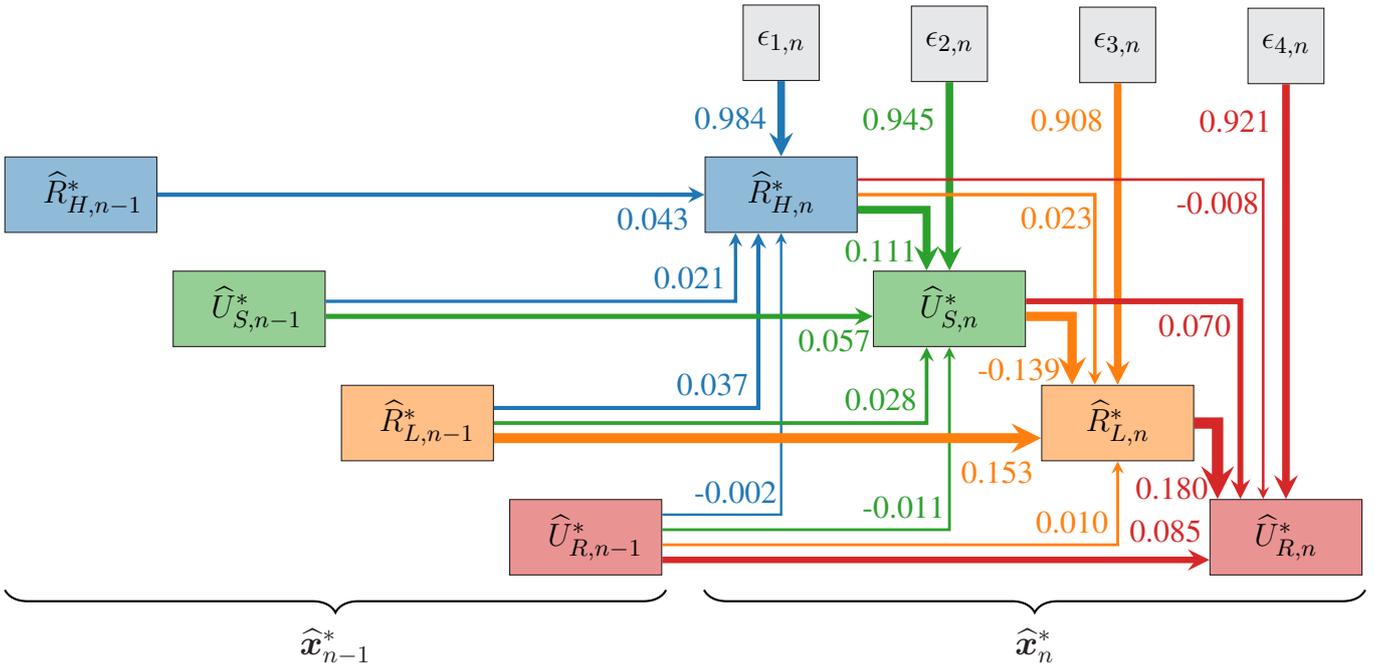

We adopt in particular a Structural VAR (SVAR) formulation of the model, which
is a factorization that makes the relationships between the contemporaneous
(same index) variables explicit. The model has the form
\begin{equation}\label{eq:SVAR1} \bm{A} \xhat^*_n = \sum_{i=1}^p \bm{C}_i
\xhat^*_{n-i} + \bm{B}\eps_n, 
\end{equation}
where $\bm{A}$, $\bm{B}$, and $\bm{C}_i$ are $4 \times 4$ matrices of model parameters, 
and $\eps_n$ is a 4-dimensional standard white noise process. With this
formulation we impose a general structure of causal ordering for the generated
random variables consistent with the chronological chain of measurement events.
Within this structure, each variable may have a causal and deterministic effect
on all future variables within range $p$, as visualized by the graph of
Fig.~\ref{fig:structural_flowchart}. The size of these effects are all subject
to fitting via the coefficients of the model. Constraints on the structural
parameters,
\begin{equation}
\bm{A} = 
\begin{bmatrix}
1 & 0 & 0 & 0 \\ 
A_{21} & 1 & 0 & 0 \\ 
A_{31} & A_{32} & 1 & 0\\ 
A_{41} & A_{42} & A_{43} & 1
\end{bmatrix}, 
\bm{B} = 
\begin{bmatrix}
B_{11} & 0 & 0 & 0 \\ 
0 & B_{22} & 0 & 0 \\ 
0 & 0 & B_{33} & 0\\ 
0 & 0 & 0 & B_{44}
\end{bmatrix}
\end{equation}
enforce the desired causal structure while assuming an uncorrelated noise
driving process. Model fitting was performed using the Python statsmodels
package \cite{seabold_statsmodels_2010}, wherein a VAR($p$) model is first fit
by ordinary least squares regression, and a maximum likelihood estimate is then
used to determine the structural decomposition.

\bigskip
\subsubsection{Device-to-device (DtD) variations}

So far, we have only considered the statistical modelling of the cycling process
of a single memory cell. However, the purpose of the presented model is to
simultaneously simulate a large number of cells in a network. Individual memory
devices on a wafer generally show statistical variations, mainly arising due to
defects and non-uniformities in fabrication \cite{fantini_intrinsic_2013,
dalgaty_situ_2021}. These DtD variations depend strongly on
the particular lithography processes and materials used. They can also originate
from intrinsic factors and are influenced by conditions during the
electroforming of each cell \cite{butcher_hot_2012, zhao_multi-level_2014}.
Because of the potential positive or negative impact on network performance, it
is important for the model to account for the DtD variability
\cite{dalgaty_situ_2021, moon_temporal_2019}.

The electrical effect of device variability is modelled with each cell using a
modification of the same underlying SVAR cycling process. Device-specific
processes are defined as members of a parametric family of processes, all based
on element-wise scaling of $\x^*_n$, where the scaling factors are themselves
random vectors. The specific process is denoted

\begin{equation}
\bm{y}_{m,n}^* = \bm{s}_m \odot \x_n^*,
\end{equation}
where $m = \{1, 2, \ldots, M\}$ is the device index, $\odot$ is the Hadamard
(element-wise) product, and $\bm{s}_m$ are $4 \times 1$ random vectors drawn from a
fixed distribution at cell initialization.

The distribution of $\bm{s}_m$ is chosen so that the features of the median
cycles of different devices are distributed and correlated in the same way as
the measured cycling data $\x_n$. This choice reflects that the covariations of
switching features DtD arise in the same physical system with causes and effects
that are comparable to those of the CtC variations. To this end, random vectors
$\widehat{\bm{s}}_m$ are drawn from a multivariate normal (MVN) distribution and
$\Gamma\inv$ is then reused to map them to the measured CtC distribution,
\begin{equation}
\bm{s}_m = \G\inv(\widehat{\bm{s}}_m) \oslash \G\inv(\bm{0}), \text{ where } \widehat{\bm{s}}_m \sim \N(\bm{0}, a \bm{\Sigma}).
\end{equation}
Here, the denominator of the Hadamard division ($\oslash$) sets the median scale vector
to the identity, ${\bm{\Sigma}=\text{cov}(\xhat_n)}$ is the sample covariance of
the normalized measurement data, and $a$ is a free scalar parameter providing
adaptability to different DtD covariance levels. A robust determination of $a$
requires measurement of many switching cycles across a large number of devices
of interest. Values in the range $a \in [1, 1.5]$ approximately correspond to published
DtD measurement samples \cite{fantini_intrinsic_2013, dalgaty_situ_2021}, but
improved processing and electroforming procedures may justify the use of $a < 1$.

\bigskip
\subsubsection{Control logic}

\begin{figure}[h]
\begin{center}
  \includegraphics[]{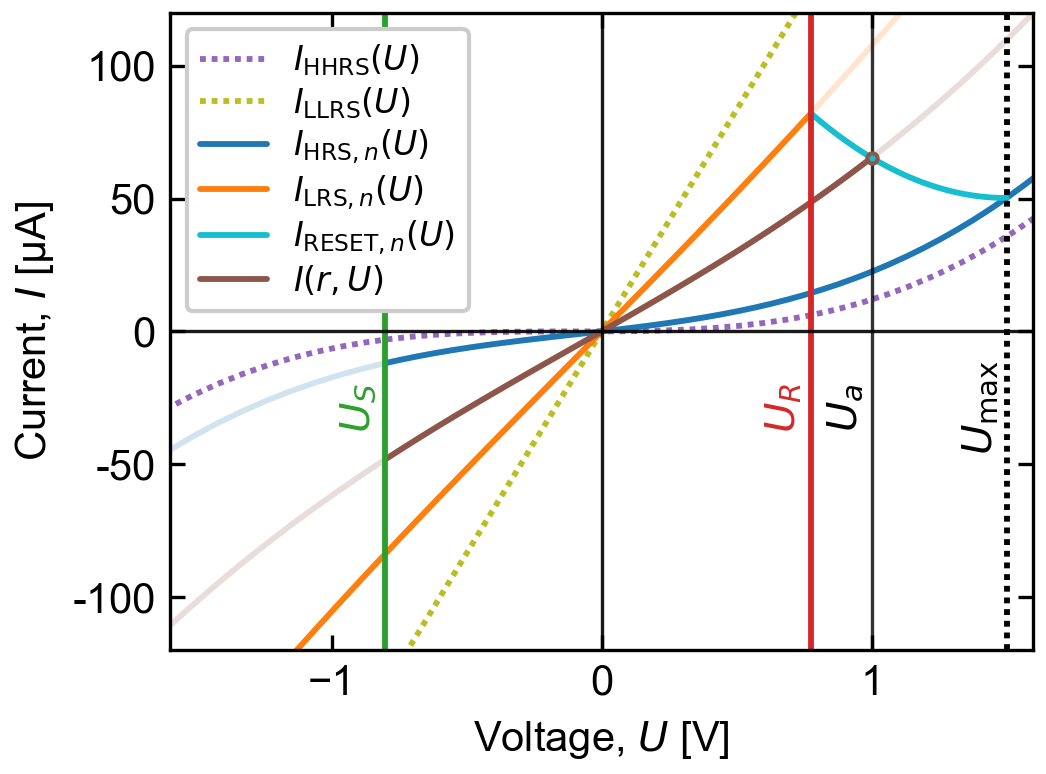}
\end{center}
\caption{
  Conduction polynomials and threshold voltages allow reconstruction of ($I,U$)
cycles from generated feature vectors. Simulated resistance switching is such
that the conduction state $I(r,U)$ induced by an applied voltage $U_a$
intersects the reconstructed cycle at $U=U_a$. For visual simplicity, the cycle
shown begins and ends in the same HRS ($R_{H,n} = R_{H,n+1}$).}
\label{fig:state_diagram}
\end{figure}

As components of a network, each simulated cell possesses a resistance state
that encodes the weight of a connection. Voltage pulses directly applied to the
cells are used to produce resistance state transitions to update the weights. In
this model, applied voltage pulses are distinguished only by a scalar amplitude
$U_a$, whether they are in fact square waveforms or they have a more complex
shape of an action potential. Although ReRAMs are known to be highly
time-dependent devices \cite{menzel_physics_2015}, we assume here that the
duration of the pulses is appropriately matched to the experimental timescale,
such that a simulated voltage pulse of a given amplitude produces an effect
comparable to the experimental voltage sweep at the instant it reaches that same
amplitude. Possible state modifications in response to an input pulse is
computed with respect to $I, U$ sweeps that are reconstructed from each
stochastic feature vector generated for each cycle as illustrated in
Fig~\ref{fig:state_diagram}.

As previously specified in eq.~\ref{eq:poly_mixture}, every possible electrical
state of a device is assumed to correspond to a polynomial $I(U)$ dependence
parameterized by a state variable $r$. It is straightforward to calculate
that the state variable for a curve passing through an arbitrary
($I,U$) point is uniquely given by the function
\begin{equation}\label{eq:r}
r(I, U) = \frac{I_{\text{LLRS}}(U) - I}{I_{\text{LLRS}}(U) - I_{\text{HHRS}}(U)}.
\end{equation}
Therefore the state variable corresponding to any static resistance level $R$
(evaluated at $U_0$) can be calculated using
\begin{equation}\label{eq:rR}
r(R) = \frac{I_{\text{LLRS}}(U_0) - U_0R\inv}{I_{\text{LLRS}}(U_0) - I_{\text{HHRS}}(U_0)}.
\end{equation}
The $I(U)$ curves for the electrical states corresponding to each cycle's HRS
and LRS, hereafter called $\IHRSn(U)$ and $\ILRSn(U)$, are defined according to
equations (\ref{eq:poly_mixture}) and (\ref{eq:rR}) such that their static
resistance equals the respective value of $R^*_{H,n}$ and $R^*_{L,n}$.

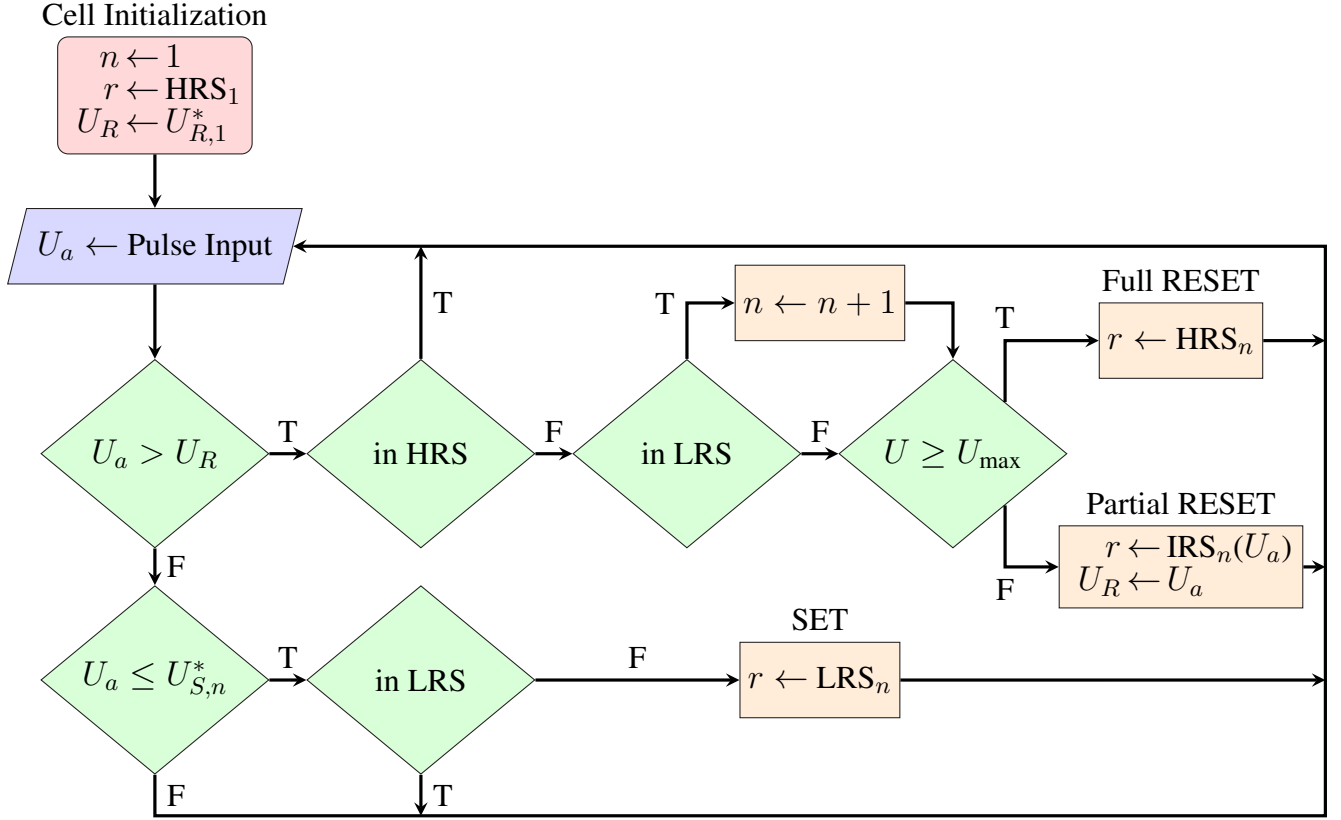
\begin{figure}[h]
\begin{center}
\tikzstyle{startstop} = [rectangle, rounded corners, minimum width=2cm, minimum height=1cm,text centered, inner sep=3pt, draw=black, fill=red!15]
\tikzstyle{io} = [trapezium, trapezium stretches=true, trapezium left angle=70, trapezium right angle=110, minimum width=2cm, minimum height=1cm, text centered, draw=black, fill=blue!15]
\tikzstyle{process} = [rectangle, minimum width=1.5cm, minimum height=1cm, text centered, inner sep=3pt, draw=black, fill=orange!15]
\tikzstyle{decision} = [diamond, inner sep=0pt, minimum width=3cm, minimum height=2.5cm, text centered, draw=black, fill=green!15]
\tikzstyle{arrow} = [very thick,->,>=stealth]


\begin{tikzpicture}[node distance=2cm, scale=.8]


  \node (start) [startstop, label={Cell Initialization}]{
    \renewcommand{\tabcolsep}{1pt}
    \begin{tabular}{rcl}
    $n$ & $\leftarrow$ & $1$ \\
    $r$ & $\leftarrow$ & HRS$_1$ \\
    $U_R$ & $\leftarrow$ & $U^*_{R,1}$
    \end{tabular}
};
\node (in1) [io, below of=start] {$U_a \leftarrow$ Pulse Input};
\node (dec1) [decision, below of=in1, yshift=-0.75cm] {$U_a > U_R$};
\node (dec2) [decision, below of=dec1, yshift=-1cm] {$U_a \leq U^*_{S,n}$};
\node (dec3) [decision, right of=dec1, xshift=1.5cm] {in HRS};
\node (dec4) [decision, right of=dec3, xshift=1.5cm] {in LRS};
\node (dec5) [decision, right of=dec4, xshift=1.5cm] {$U \geq U_\text{max}$};
\node (dec6) [decision, right of=dec2, xshift=1.5cm, yshift=0cm] {in LRS};
\node (pro1) [process, right of=dec5, xshift=1cm, yshift=1.5cm, label={Full RESET}] {$r \leftarrow$ HRS$_n$};
\node (pro2) [process, right of=dec5, xshift=1cm, yshift=-1.5cm, label={Partial RESET}] {
    \renewcommand{\tabcolsep}{1pt}
    \begin{tabular}{rcl}
    $r$ & $\leftarrow$ & IRS$_n$($U_a$) \\
    $U_R$ & $\leftarrow$ & $ U_a$
    \end{tabular}
    };
\node (pro3) [process, right of=dec6, xshift=3.25cm, label={SET}] {$r \leftarrow$  LRS$_n$ };
\node (pro4) [process, above of=pro3, yshift=3cm] {$n \leftarrow n+1$};

\draw [arrow] (start) -- (in1);
\draw [arrow] (in1) -- (dec1);
\draw [arrow] (dec1) -- node[anchor=west] {F} (dec2);
\draw [arrow] (dec1) -- node[anchor=south] {T} (dec3);
\draw [arrow] (dec2) -- node[anchor=south] {T} (dec6);
\draw [arrow] (dec3) -- node[anchor=south] {F} (dec4);
\draw [arrow] (dec4) -- node[anchor=south] {F} (dec5);
\draw [arrow] (dec4) |- node[anchor=east] {T} (pro4);
\draw [arrow] (pro4.0) -| node[anchor=south] {} (dec5.90);
\draw [arrow] (dec5.45) |- node[anchor=south] {T} (pro1);
\draw [arrow] (dec5.315) |- node[anchor=north] {F} (pro2);

\coordinate (SE) at ($(dec2) + (19.25, -2.25)$);
\draw [arrow] (dec2) -- node[anchor=west] {F} (dec2 |- SE) -- (SE) |- (in1);

\draw [arrow] (dec6) -- node[anchor=south] {F} (pro3);
\draw [arrow] (dec6) -- node[anchor=west] {T} (dec6 |- SE);
\draw [arrow] (dec3) -- node[anchor=west] {T} (dec3 |- in1);
\draw [arrow] (pro1) -- (pro1 -| SE);
\draw [arrow] (pro2) -- (pro2 -| SE);
\draw [arrow] (pro3) -- (pro3 -| SE);


\end{tikzpicture}
\caption{Logical flow chart showing how applied voltage pulses affect the
state of each cell during simulation. Following the experimental observations,
SET processes always occur abuptly below a threshold voltage, while partial
switching is induced for a range of RESET voltages, with intermediate states
bounded for cycle $n$ by resistance values between $R_{L,n}$ and $R_{H,n+1}$. As
resistance cycling progresses, later terms of the stochastic driving process are
used for limiting resistance states and threshold voltages. Pulse amplitudes not
producing a state change are efficiently disregarded.
}\label{fig:pulsing_flowchart}
\end{center}
\end{figure}

Transitions between HRS and LRS states in response to an applied pulse amplitude
$U_a$ follow an empirically motivated structure, represented by the flow chart
of Fig.~\ref{fig:pulsing_flowchart}. The SET transition for the $n$th cycle
HRS$_n$ $\rightarrow$ LRS$_n$ may occur for negative voltage polarities and
follows a simple threshold behavior, fully and instantaneously transitioning the
first time a voltage pulse with amplitude $U_a \leq U^*_{S,n}$ is applied. In
contrast, the RESET transition LRS$_n$ $\rightarrow$ HRS$_{n+1}$ occurs
gradually in the positive polarity with increasing $U_a$ in the range $U^*_{R,n}
< U_a \leq \Umax$, where $\Umax=1.5$~V is the maximum voltage applied in the
voltage sweeping measurement. A transition curve $I_\text{RESET,n}(U)$ is
defined to connect the ($I,U$) points of the two limiting states where the
RESET transition begins and ends. The functional form of the transition curve is
chosen to be the parabola with boundary conditions
\begin{eqnarray}
I_{\text{RESET,n}}(U^*_{R,n}) &=& I_{\text{LRS,n}}(U^*_{R,n}) \\
I_{\text{RESET,n}}(\Umax) &=& I_{\text{HRS,n+1}}(\Umax) \\
\left. \frac{dI_\text{RESET,n}}{dU} \right|_{U=\Umax} &=& 0.
\end{eqnarray}
When a voltage pulse in the RESET range is applied, an intermediate resistance
state (IRS) results which is calculated with reference to the transition curve
such that $I(r, U_a) = I_\text{RESET,n}(U_a)$. Additional RESET pulses with
larger amplitudes may be applied to incrementally increase the cell resistance,
with HRS$_{n+1}$ being reached only if $U_a \geq \Umax$, after which no further
RESET switching is possible for the $n$th cycle. After either partial or full
RESET, the resistance may only decrease again by entering the following
LRS$_{n+1}$ with a voltage pulse meeting the SET criterion $U_a \leq
U^*_{S,n+1}$.

\bigskip
\subsubsection{Readout}
Simulated current measurements (readouts) for each individual cell can be generated
given an arbitrary readout voltage input $U_\text{read}$. The noise-free current
level simply corresponds to evaluation of $I(r, U_\text{read})$ for each cell.
In any real system, however, current readouts are accompanied by measurement
noise, which may impact system performance and even present a fundamental
bottleneck. Furthermore, in digital systems current readouts are converted to
finite resolution by analog to digital converters (ADCs). Due to constraints of
power consumption and chip area, ADC resolution is often limited such that
digitization is the dominant contributor to the total noise
\cite{ma_non-volatile_2019}. Many additional noise sources can be considered,
such as $1/f$ noise \cite{wiefels_hrs_2020}, but at minimum the Johnson-Nyquist noise
and the shot noise should be included because they represent a lower bound of noise
amplitude impacting all systems.

To account for measurement noise, each individual current readout includes an
additive noise contribution drawn from a normal distribution. The noise
amplitude is approximated from the Nyquist and Schottky formulas,
\begin{equation}
\sigma_{I} = \sqrt{\frac{4 k_B T I_\text{read} \Delta f}{U_\text{read}} +  2 q I_\text{read} \Delta f},
\end{equation}
where $\Delta f$ is the noise equivalent bandwidth, $k_B$ is the Boltzmann constant,
$T=300$~K is the temperature, $q$ is the electron charge, $I_\text{read}$ is the noiseless
current readout, and $U_{\text{read}}$ is the voltage used for readout. The
total current is then ideally digitized with an adjustable resolution
$n_{\text{bits}}$ between adjustable minimum $I_\text{min}$ and maximum
$I_\text{max}$ current levels.

\bigskip
\subsection{Program implementation}

To facilitate investigations of neuromorphic systems, model implementations
designed to simulate arrays of devices were developed in
the Julia programming language. Julia is a modern high-level language that is
focused on performance and that provides an advanced ML and
scientific computing ecosystem. Julia programs compile to efficient native code
for many platforms via the LLVM compiler infrastructure, and a cursory analysis
indicated that single threaded CPU performance of a Julia implementation is up
to 5,000 times faster than a Python implementation. Furthermore, as modern computational
resources are highly parallel, Julia's support for CPU multi-threading and GPU
programming through CUDA.jl \cite{besard_effective_2019} is an important advantage.

All model parameters corresponding to the device characterized in this article,
including different possible SVAR model orders, $p \in [1, 200]$, are stored in
a binary file which is read in by the program at startup. Each instantiated cell
stores state information and $p$ cycles of history using primarily 32-bit
floating point numbers. The total memory footprint grows linearly with the
chosen model order and is approximately $16p + 56$~bytes per cell. A
reduced form VAR process is used to compute realizations of
$\x^*_n$, which are lazily evaluated along with the parabolic transition
polynomials if and when they are needed. The majority of the necessary runtime
computations are formulated as matrix multiplications,
which are heavily optimized operations across many different contexts.

The present release contains two model implementations in order to suit a wide
variety of computing platforms and use cases \cite{t_hennen_2022_5881893}. The
first is a CPU optimized version wherein the cells of an array are individually
addressable for read/write operations. These operations are naturally
parallelized for multi-core processors by partitioning the cells and assigning
each partition to independent threads of execution. The second implementation is
a GPU accelerated version compatible with CUDA capable GPUs. This version uses a
vectorized data structure and parallel array abstractions to take advantage of
the implicit parallelism programming model of CUDA.jl. Here, all defined cells
are always accessed simultaneously, with each read/write operation employing
optimized linear algebra GPU kernels. While the GPU implementation integrates
well with other ML components risiding in GPU shared memory and achieves higher
throughput per cell for large parallel operations, the CPU implementation
obtains higher update rates for sparse operations commonly
encountered in large-scale models
\cite{pedroni_memory-efficient_2019, pedroni_design_2020}.

\bigskip
\section{Results} 

\begin{figure}[h]
\begin{center}
\includegraphics{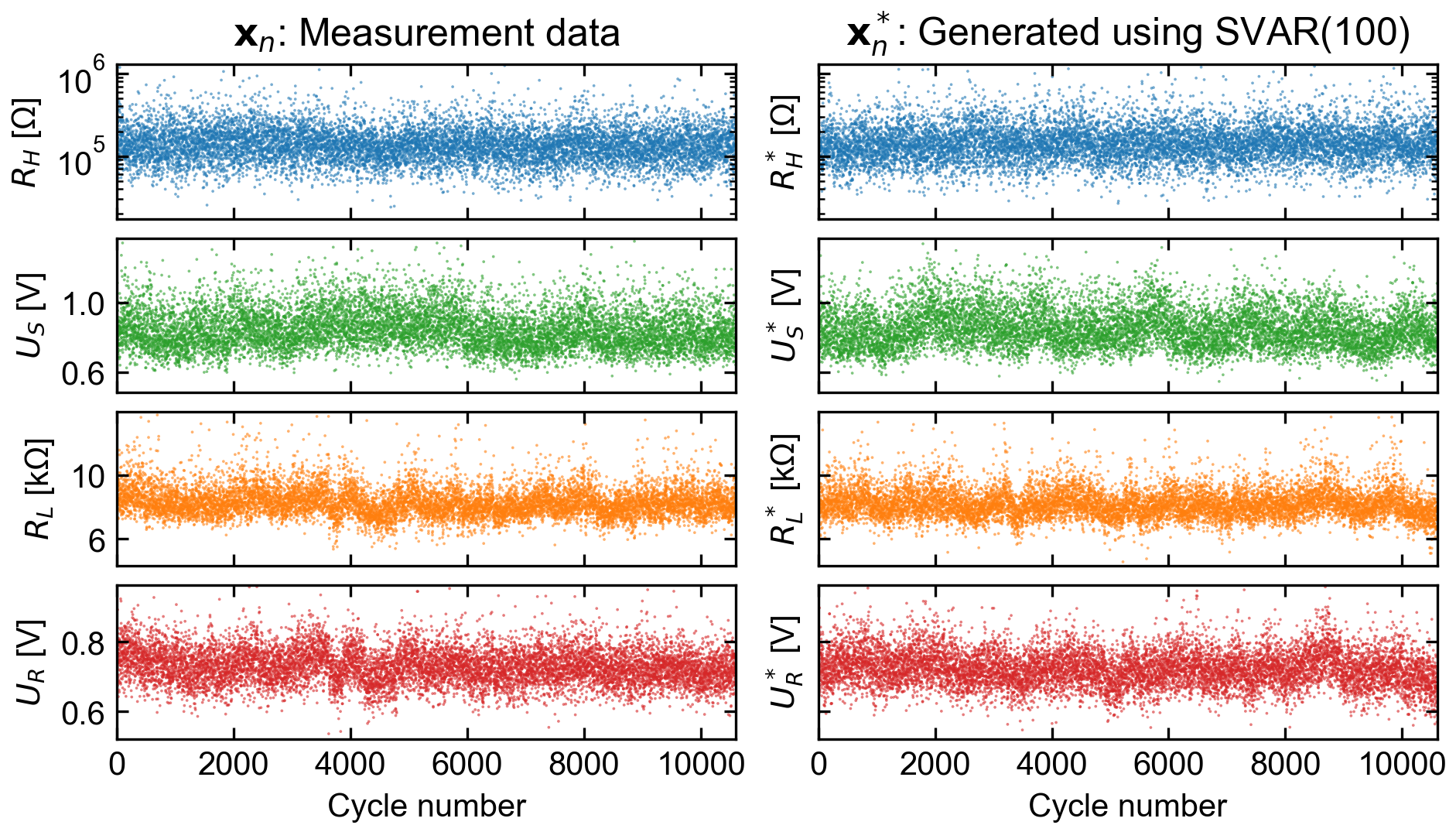}
\end{center}
\caption{Comparison of feature time series extracted from measurement data and
  those generated by the SVAR-based model. The compared features converge to
  effectively equivalent distributions and the short-range behavior is qualitatively
  similar across thousands of cycles.}\label{fig:scatter_comparison}
\end{figure}

As shown visually in the scatterplot of Fig.~\ref{fig:scatter_comparison},
the stochastic process $\x^*_n$ generates data that closely resemble the
measurement data $\x_n$. To quantify the difference between the generated
distributions and the empirical distributions, the first Wasserstein distance \cite{kantorovich_mathematical_1960}
was calculated element-wise and averaged across 100 realizations of $\x_n^*$
with length $10^6$. The result,
\begin{equation}
\label{eq:wasserstein}
\overbar{\bm{W}}_1(\x_n, \x_n^*) = 
\begin{bmatrix}
5146~\Omega\\ 
937~\upmu\text{V}\\ 
20~\Omega\\ 
356~\upmu\text{V}
\end{bmatrix},
\end{equation}
is much smaller than the mean feature vector, $\bar{\x}_n$ (Eq. \ref{eq:xbar}),
and independent of the chosen model order. This shows that the goal of
reproducing the measurement distributions is well achieved for the input dataset
by using the described method of probability density transformation.

\begin{figure}[h]
\begin{center}
  \includegraphics[]{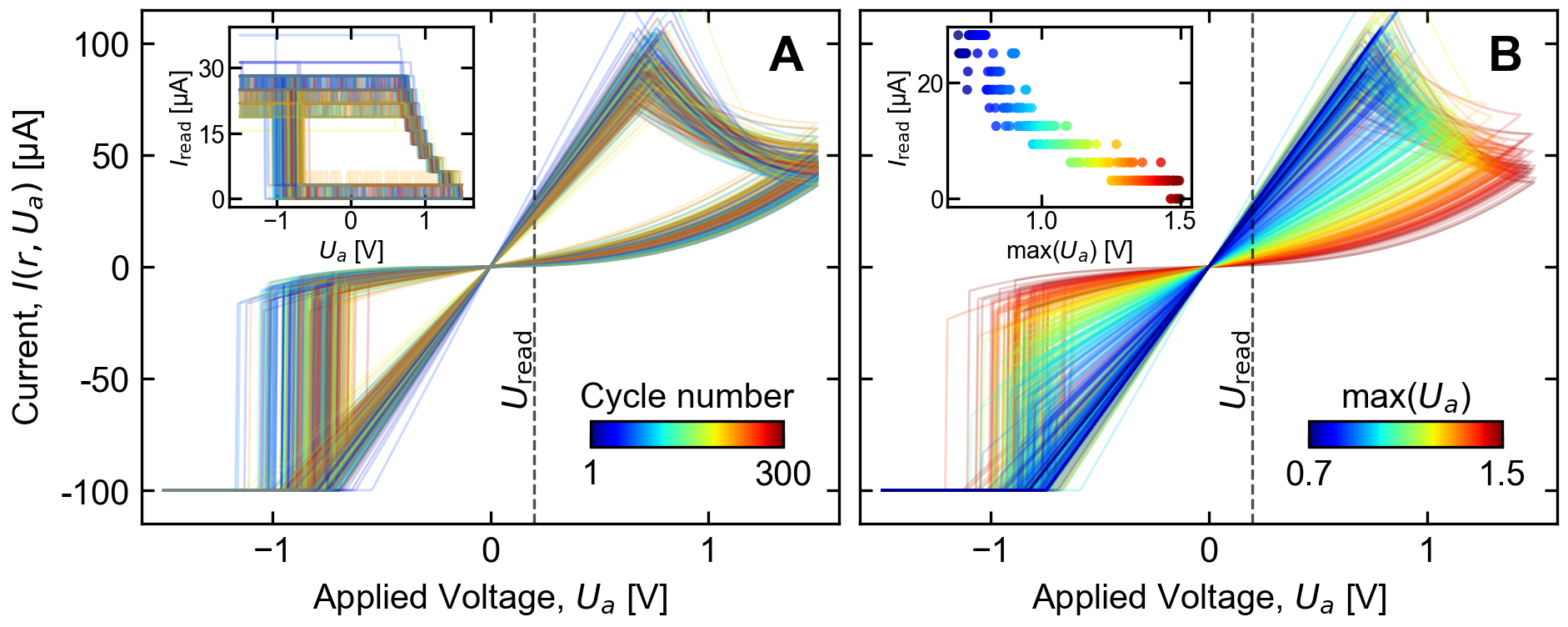}
\end{center}
\caption{
  Two example simulations involving repeated cycling of a single device. Voltage
pulse sequences were applied with varying amplitude following a triangular
envelope, and the ($I,U$) characteristic of each cycle is plotted in a
different color. Subplot \textbf{(A)} shows 300 consecutive cycles between the
full voltage range $\pm$1.5~V, with a readout performed after every pulse
(inset). Subplot \textbf{(B)} demonstrates multilevel capability with 300 cycles between
-1.5~V and maximum voltage that increases each cycle, from 0.7~V to 1.5~V.
Readouts following each cycle are shown in the inset. In each case, readouts
were simulated using a fixed $U_\text{read}=$~200~mV, including noise and 4-bit
quantization between $I_\text{min}= 0~\upmu$A and
$I_\text{max}=40~\upmu$A.} \label{fig:simulated_loops}
\end{figure}

Simulations of full ($I,U$) cycling measurements
[Fig.~\ref{fig:simulated_loops}(A)] show close similarity 
with the measurement data of Fig.~\ref{fig:ivloops_measured}.
Multi-resistance-level capability is also demonstrated by a similar simulation
involving partial RESET operations by changing the maximum voltage applied
[Fig.~\ref{fig:simulated_loops}(B)]. The dependence of the resulting HRS value
on the applied voltage reproduces a non-linear characteristic comparable to
experimental findings \cite{park_nanoscale_2013, ambrogio_neuromorphic_2016}.

\begin{figure}[h]
\begin{center}
  \includegraphics[]{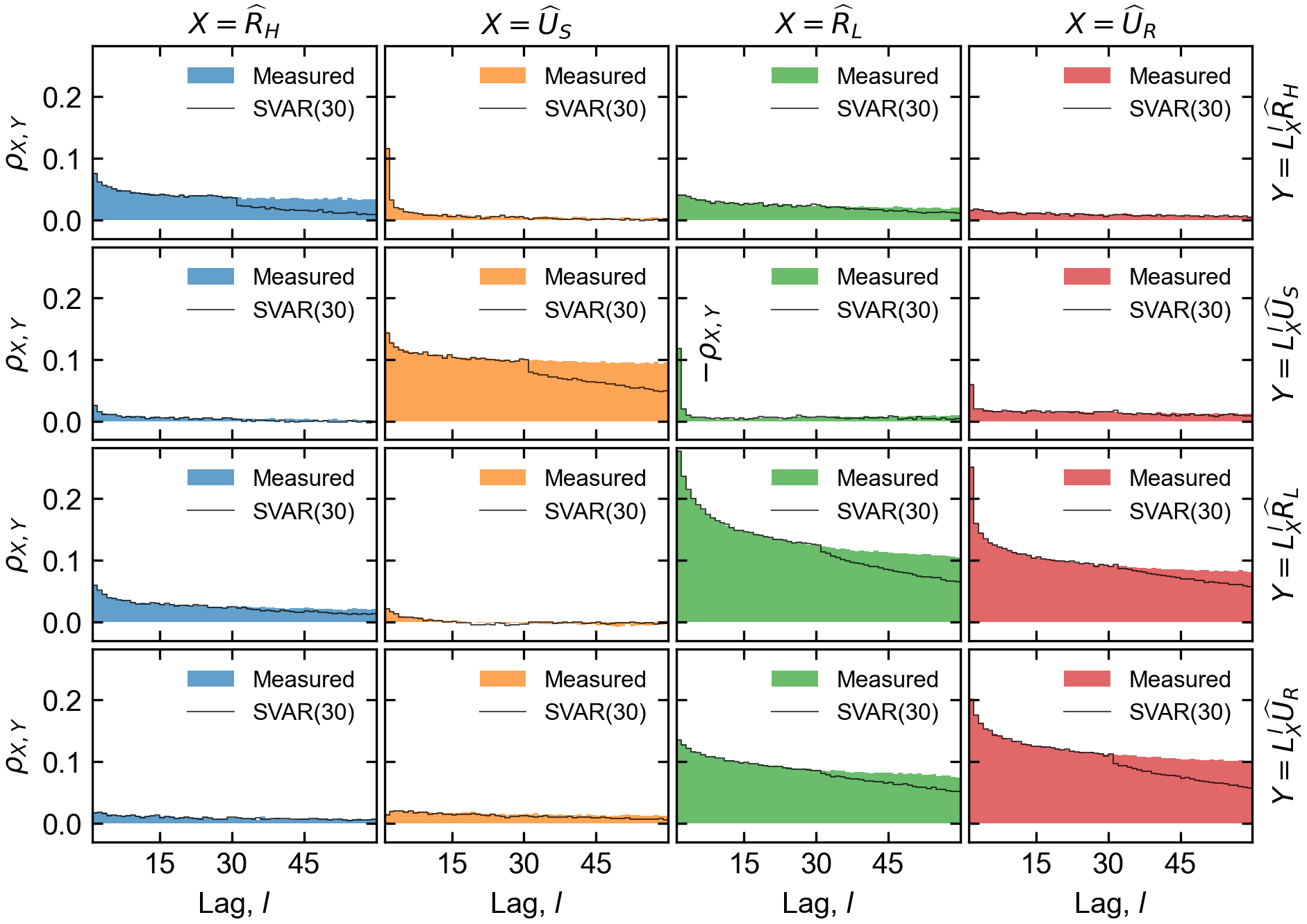}
\end{center}
\caption{
  Auto- and cross-correlations of the normalized feature vector
components, showing the pearson coefficients $\rho_{X,Y}$ of the variables
specified in the subplot columns $X$ and rows $Y$ as a function of lag $l$. Row
variables are lagged with respect to column variables, as denoted by the lag
operators $L_X$. A comparison between measurement data and data generated from
SVAR(30) shows extremely close agreement up to cycle range 30. For lags larger
than the chosen model order, some of the correlations of $\xhat^*$ decay more
quickly than $\xhat$.
}\label{fig:autocorrelation}
\end{figure}

While a full structural analysis of the fitted SVAR($p$) model parameters
($\bm{A}, \bm{B}, \bm{C}_i$) will not be
presented here, a few aspects
are worthy of note. For the fit corresponding to the
particular device and measurement described here, the white noise terms are by
far the dominant contributors to all four modelled features. The contemporaneous
terms ($\bm{A}$) and first order ($\bm{C}_1$) terms are the next most
significant, which indicates that the most recent cell history is most relevant
for generating the proceeding states. Nevertheless, input data correlations
persist for many cycles, and the generating process $\x^*_n$ successfully
reproduces the overall correlation structure of the data up to at least $p$
cycle lags, as shown in detail in Fig.~\ref{fig:autocorrelation}.

Although no physical effects were explicitly put into the model definition, it
is important to recognize that they are quantitatively captured and put into a
useful statistical context by the SVAR model fitting procedure. As seen in the
graph of Fig.~\ref{fig:structural_flowchart}, the strongest deterministic
coefficients in the fitted model correspond to the relationships
\begin{align}
\widehat{R}^*_{H,i}   & \xrightarrow{\phantom{-}0.111}    \widehat{U}^*_{S,i} ,\label{eq:c1}\\
\widehat{U}^*_{S,i}   & \xrightarrow{-0.139}              \widehat{R}^*_{L,i} ,\label{eq:c2}\\
\widehat{R}^*_{L,i-1} & \xrightarrow{\phantom{-}0.153}    \widehat{R}^*_{L,i} ,\label{eq:c3}\\
\widehat{R}^*_{L,i}   & \xrightarrow{\phantom{-}0.180}    \widehat{U}^*_{R,i} .\label{eq:c4}
\end{align}

Comparable relationships between switching variables have been identified and
discussed in physics-based models and simulations as well as in experimental
studies involving various materials \cite{ielmini_modeling_2011, nardi_control_2011,
  nardi_resistive_2012, nishi_effect_2015, kim_impact_2016,
  la_torre_dependence_2016, kim_voltage_2016}. According to relation \ref{eq:c1},
larger starting HRS values tend to contribute to a higher SET voltage, which is
a well known effect due to a reduced driving force for ionic motion at a given
applied voltage, as a larger HRS gives both reduced power dissipation as well as
a reduced electric field in a thicker insulating gap. The subsequent LRS is
strongly affected by the SET voltage (relation \ref{eq:c2}). This can be
attributed to the runaway nature of the SET
transition and a higher voltage initial condition, and is also connected with
the dynamics of the current limiting
circuitry \cite{hennen_current-limiting_2021}. The LRS value is also strongly
correlated with the value of the previous LRS (relation \ref{eq:c3}), because of
the influence of the residual filamentary structure from the previous
cycle \cite{piccolboni_investigation_2015}. Lastly, relation \ref{eq:c4}
indicates that higher LRS values tend to have larger reset voltages, which has
to do with a balance of factors influencing filament dissolution, including
temperature and drift. This balance depends on the cell materials, operating
regime, and internal series resistance \cite{ielmini_universal_2011}.

\bigskip
\subsection{Benchmarks}

As a benchmark of the throughput of write operations, repeated resistance
cycling was induced on arrays of simulated cells under varying conditions. In
each case, voltage pulse sequences to be applied to all defined cells were
generated prior to the benchmarks, consisting of amplitudes
$\pm$1.5~V with alternating polarity. Defined as such, every pulse drives
each cell through a transition into its next HRS or LRS. The read operation
was benchmarked separately under equivalent conditions, reading out the entire
array using a fixed readout voltage of $U_{\text{read}} = 0.2~V$.

\begin{figure}[h]
\begin{center}
  \includegraphics[]{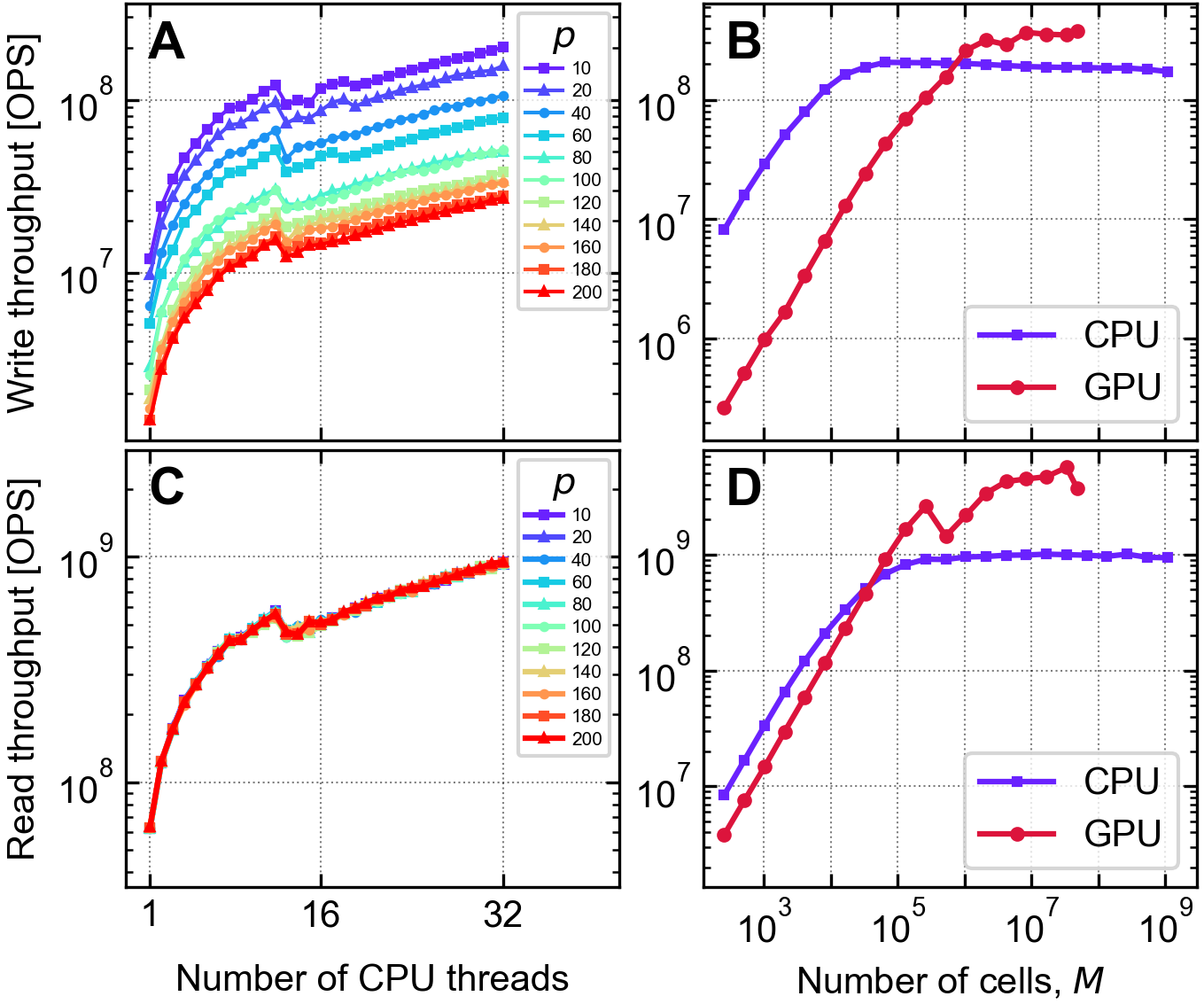}
\end{center}
\caption{
  Benchmarks of the read/write operation throughput per cell of the Julia model
implementations. In \textbf{(A)} and \textbf{(C)}, an array of $2^{20}$
(${\sim}10^6$) cells are simulated on the CPU as a function of number of parallel
threads spawned, and the VAR model order $p$. In \textbf{(B)} and \textbf{(D)},
the CPU (32 threads) and GPU implementations are benchmarked versus the cell array
size $M$, with $p = 10$.
}\label{fig:benchmark}

\end{figure}

The CPU benchmark was performed using an Intel Xeon Silver 4116 CPU, varying
the cell array size $M$, the order of the VAR process $p$, as well as the number
of threads used to perform the operations in parallel. The resulting read/write
throughputs are summarized in Fig.~\ref{fig:benchmark}. Write throughputs up to
$2\times 10^8$ operations per second (OPS) were obtained, which is equivalent to
5~ns per individual write operation. Read operations were nearly an order of
magnitude faster than writes, with up to $10^9$ OPS or 1~ns per read operation.
Due to the size of necessary matrix multiplications, increasing the VAR order
$p$ incurs a cost of write throughput, with a $p=100$ model running
approximately 4$\times$ slower than one with $p=10$. The read operation, in
contrast, shows a negligible dependence on the VAR order.

The GPU accelerated version was benchmarked in an analogous way, using the same
host machine with an NVIDIA TITAN RTX GPU device. The results are shown in
dependence of the cell array size $M$ in Fig.~\ref{fig:benchmark}(B,D). The GPU
implementation overtakes the CPU above $M=10^6$ parallel operations where the
entire array is updated, and achieves 2$\times$ faster updates and 5$\times$
faster readouts for large arrays with $M > 10^7$. However, CPU throughput is
applicable to subsets of the array, and may retain an advantage for sparse
operations.

\bigskip
\section{Discussion} 

In order to assess the potential of emerging synaptic
devices, new lightweight and accurate device models are needed to constitute
the millions/billions of weights used in modern machine learning (ML) models. Candidate memory cells
such as ReRAM are highly non-linear stochastic devices with complex internal
states and history dependence, all of which needs to be
explicitly taken into account. In this article we introduced an efficient
generative model for large synaptic arrays, which closely reproduces the
statistical behavior of real devices.

Taking advantage of a recently developed electrical measurement
technique \cite{hennen_current-limiting_2021}, we
systematically fit the model to a dataset that is
dense in relevant information about the device state
evolution. Together with this new kind of measurement, our modelling approach
helps complete a neuromorphic design feedback loop by defining
a programmatic connection from the measured behavior of a fabricated device
under the intended operating conditions directly to fitted model parameters.
Probability density transformation of the underlying SVAR stochastic process
gives the model the power to accurately reproduce nearly arbitrary distribution
shapes and covariance structures across the switching cycles and across the
separate devices. These features enable evaluation of network performance while
automatically adapting to a wide variety of possible future device designs.

We provide parallelized implementations for both CPU and GPU, where up to
15 million cells per GB of available memory can be simulated at once.
Benchmarks show throughputs above three hundred million weight updates per
second, which exceeds the pixel rate of a 30 frames per second video stream at 4K
resolution (3840$\times$2160 pixels). Realistic current readouts including
digitization and noise were also benchmarked, and are approximately an order of
magnitude faster than weight updates. While speeds can be expected to improve
with future optimizations, these benchmarks give a basis for
estimating the scope of applicability of the model to ML
tasks.

The implementation and the general concept of this model are naturally
extendable. Although model parameters were adapted here to a specific
HfO$_2$-based ReRAM device, the method is applicable to a variety of other types
of stochastic memory cells such as PCM, MRAM, etc. Four specific switching
features were chosen in this demonstration to reconstruct ($I,U$) cycling
behavior, but additional switching parameters can also be extracted from
measurements and accommodated within this framework. Ideally informed by
statistical measurement data, different functional forms, transition behaviors,
time dependence, and underlying stochastic processes can each be substituted.
Fitting may also be performed with respect to the output of physics-based
simulations, thereby establishing an indirect link to physical parameters while
achieving much higher computational speed. With these considerations, the model
represents a flexible foundation for implementing large-scale neuromorphic
simulations that incorporate realistic device behavior.






\section*{Conflict of Interest Statement}
The authors declare that the research was conducted in the absence of any commercial or financial relationships that could be construed as a potential conflict of interest.

\section*{Author Contributions}

TH wrote the manuscript, performed data analysis, and implemented the model. AE
carried out the ($I,U$) measurement. JN and GM fabricated the ReRAM devices. RW and DW co-advised the project, and DB conceived of the concept and was in charge of the project.


\section*{Acknowledgments}
The authors thank Thomas P\"{o}ssinger of RWTH Aachen for illustrating Figs \ref{fig:synapse} and \ref{fig:correlated_cartoon}.

\section*{Code Availability}
A Julia implementation of the model is available on GitHub (\url{https://github.com/thennen/StochasticSynapses.jl}) and archived in Zenodo (\url{https://doi.org/10.5281/zenodo.6535411}).

\bibliographystyle{frontiersinHLTHFPHY} 
\bibliography{stochasticReRAM}


\end{document}